%% file: main.tex
\documentclass{article} %
\usepackage{iclr2027_conference,times}

\input{math_commands.tex}

\usepackage{hyperref}
\hypersetup{colorlinks=true,linkcolor=ljsLink,citecolor=ljsLink,urlcolor=ljsLink}
\usepackage{url}

\title{Latent Jacobian Steer: Eliciting Contact-Rich\\Out-of-Distribution Behaviors from\\Pretrained Humanoid Controllers}

\author{S K Swaminathan$^{1,2,*}$ \quad Shivam Sood$^{1}$ \quad Shreyansh Kansal$^{1}$ \\
\textbf{Chengyang He$^{1}$} \quad \textbf{Guillaume Adrien Sartoretti$^{1}$} \\[4pt]
\normalfont $^{1}$National University of Singapore \quad $^{2}$Indian Institute of Technology Kharagpur \\
\normalfont \url{https://latent-jacobian-steer.github.io} \quad $^{*}$\texttt{swami@u.nus.edu}
}

\iclrfinalcopy

\usepackage{xcolor}
\usepackage{listings}
\usepackage{booktabs}
\usepackage{array}
\usepackage{longtable}
\usepackage{placeins}
\usepackage{caption}


\newcommand{\stackhead}[1]{\begin{tabular}[b]{@{}c@{}}#1\end{tabular}}

\definecolor{ljsLink}{RGB}{20,70,150}
\definecolor{ljsCodeBackground}{RGB}{247,247,247}
\definecolor{ljsCodeComment}{RGB}{0,105,45}
\definecolor{ljsCodeString}{RGB}{130,45,145}

\lstdefinestyle{ljsPython}{
  language=Python,
  basicstyle=\ttfamily\scriptsize,
  keywordstyle=\color{blue!60!black},
  commentstyle=\color{ljsCodeComment},
  stringstyle=\color{ljsCodeString},
  backgroundcolor=\color{ljsCodeBackground},
  numbers=left,
  numberstyle=\tiny\color{black!50},
  numbersep=6pt,
  frame=single,
  rulecolor=\color{black!20},
  breaklines=true,
  breakatwhitespace=false,
  columns=fullflexible,
  keepspaces=true,
  showstringspaces=false,
  tabsize=4,
  captionpos=b,
  xleftmargin=1.5em,
  framexleftmargin=1.2em
}

\begin{document}

\maketitle

\begin{figure}[!b]
\centering   \centering\includegraphics[width=0.72\linewidth]{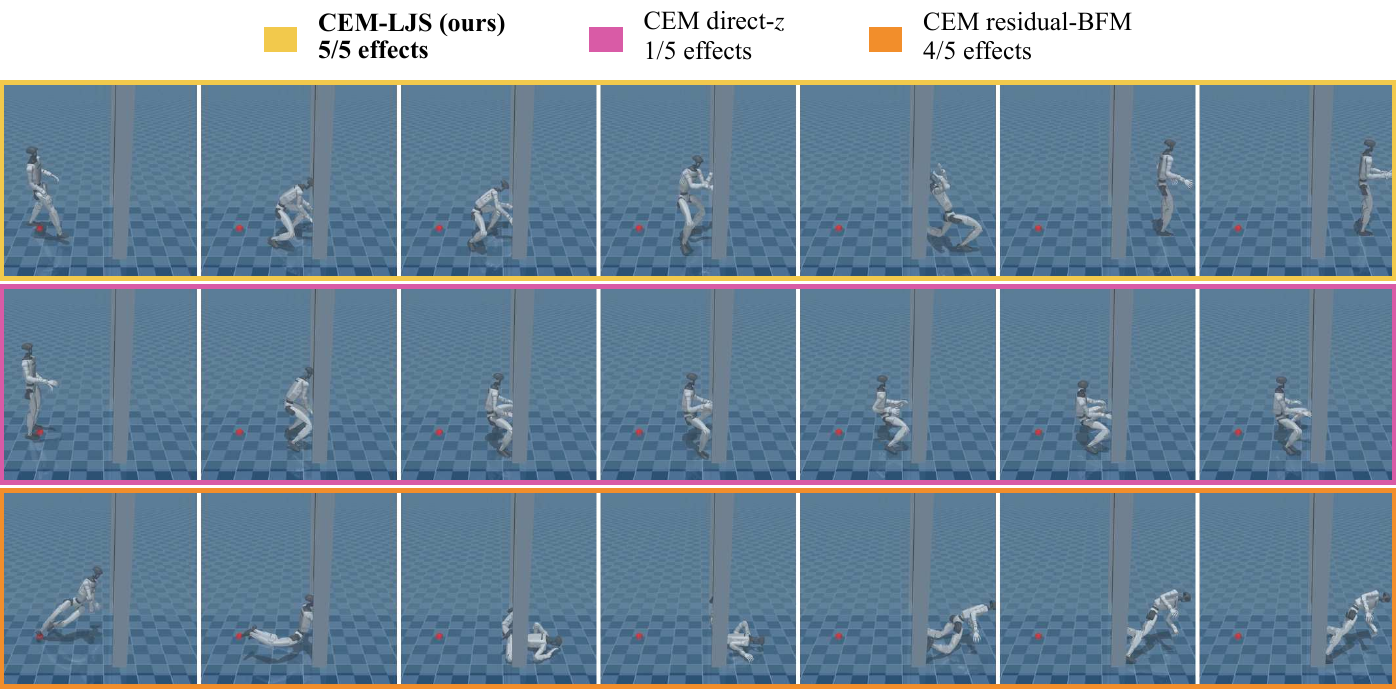}   \caption{\textbf{Shutter traversal} Chronological frames from comparing CEM-based \citep{Rubinstein1999} searches of our method and the best seeds of baseline methods {over ten seeds}. \textbf{CEM-LJS(ours)} lifts the shutter{, crosses on its feet, and stands up}. CEM with direct-$z$ search {never lifts the shutter}. CEM with residual raw joint-action search over BFM {hacks the task by crawling on its torso}.}   \label{fig:shutter_qualitative}
\end{figure}

\begin{abstract}

Given a pretrained general latent-conditioned whole-body humanoid controller{, say a behavioral foundation model (BFM),} how can we elicit complex and contact-rich out-of-distribution behaviors? {Raw} joint-action exploration enables direct{,} task-specific adjustments but can disrupt coordination learned by the pretrained controller, whereas {latent exploration retains that coordination} but is difficult to steer toward desired physical effects. To combine the complementary strengths of these exploration spaces, we propose Latent Jacobian Steer (LJS), which couples sampling in direct joint-action space with the semantics of the learned frozen controller's input latent space. For a task specified through desired physical effects, we sample task-relevant joint-action perturbations and map them through a damped local inverse of the policy Jacobian to obtain candidate latent {($z$)} commands.  These commands are executed through the frozen pretrained policy, while exploration is directed toward the physical outcomes required by the new task. Candidate outcomes are evaluated in simulation and composed into latent-command sequences that accomplish the task. Ablations show that neither action-space nor latent-space exploration \textit{alone} provides the same capability as their coupling. Across diverse out-of-distribution tasks, {simulation-based} searching using this method produces contact-rich behaviors, including hurdling, manipulating articulated objects, and opening doors through non-prehensile contact. {Over ten seeds, LJS completes more of each humanoid task than when searching in either space alone. Beyond humanoids, it also steers a goal-conditioned dexterous hand policy through four assembly tasks better than searching in goal space directly.}
\end{abstract}

\section{Introduction}

For humanoid robots to become useful, they must be able to physically interact with objects and environments in diverse and often contact-rich ways. Recent advances in motion capture and general motion-tracking have enabled humanoids to reproduce an increasingly broad repertoire of dynamic whole-body behaviors~\citep{chen2025gmt, liao2025beyondmimicmotiontrackingversatile}. Yet obtaining behaviors for a new contact-rich loco-manipulation task remains considerably more task-specific. Such behaviors often require dedicated human demonstrations or trajectories reconstructed from video through Real2Sim pipelines that recover the relevant object and environment interactions~\citep{dan2025xsim, ranawaka2026simfoundrymodularautomatedscene}.

High-fidelity simulation provides a natural way to generate such behaviors. Existing approaches using simulation either learn new behaviors from task-specific rewards~\citep{sferrazza2024humanoidbench, DrS} or explicitly generate demonstrations and trajectories for them~\citep{jing2025humanoidgen, taouil2026motiondisco}.
More recently, Behavioral Foundation Models such as Meta Motivo and BFM-Zero have instead pretrained general latent-conditioned whole-body controllers that can be prompted towards different motions, goals, and rewards~\citep{tirinzoni2025zeroshot, li2026bfmzero}. These approaches inspire our question: can we steer such a frozen controller to produce complex contact-rich behaviors that lie outside its training distribution, without collecting a task-specific motion demonstration or training a new policy?

We hypothesize that much of the whole-body coordination required for these behaviors is already present in a sufficiently general pretrained controller, even if it was never trained on such object-environment interaction data. 
This is fairly intuitive in hindsight. A human who has learned to move in free space can use these motion priors to play around with objects and the environment. The challenge is how to \textit{steer} it towards a new physical interaction. Direct exploration in joint-action space makes task-specific changes easy to express, for example, moving a hand towards a desired contact, but it can disrupt the coordination learned by the controller. Exploration in its latent space preserves this coordination, but it is difficult to know which latent perturbation will produce a desired physical effect. Our key observation is that these two spaces are locally connected by the frozen policy itself: its Jacobian describes how perturbations in latent commands change the resulting joint actions.

Based on this observation, we introduce \textbf{Latent Jacobian Steer (LJS)} (Figure~\ref{fig:ljs_overview}), which couples precise exploration in joint-action space with coherent execution through the latent-conditioned pretrained controller. Given a task-relevant joint-action perturbation, LJS uses a damped local inverse of the policy Jacobian to obtain candidate latent commands, which are then executed and evaluated in simulation. Complex tasks can then be described through simple sequences of physical effects: opening a door, for example, can be specified as (1) reaching and grasping the handle, (2) manipulating it to release the latch, and (3) pushing the door about its hinge. In this way, LJS provides a means of eliciting and collecting new contact-rich behaviors from a frozen whole-body controller without requiring a task-specific motion demonstration, a complete reference trajectory, or training a new task policy. {We evaluate LJS over ten seeds on three contact-rich humanoid tasks and, beyond humanoids, on four precise assembly tasks with a frozen goal-conditioned dexterous hand policy.}

\section{Related Works}

\textbf{From motion tracking to contact-rich behavior discovery.}
General trackers such as TWIST \citep{TWIST} and SONIC \citep{SONIC} provide broad whole-body motion repertoires, while PGMT \citep{PGMT} and GigaBrain-WBC-0.5 \citep{GigaBrainWBC} incorporate terrain perception and environment interaction. Motion generators such as OMOMO \citep{OMOMO} and CHOIS \citep{CHOIS} produce human--object interactions from object trajectories or language and spatial guidance, using captured interactions as training data. For a new contact-rich task, however, the movement that realizes the required interaction must still be obtained. Real2Sim pipelines reconstruct it from demonstrations \citep{dan2025xsim, ranawaka2026simfoundrymodularautomatedscene}; contact-implicit trajectory optimization \citep{Posa2014} and logic-geometric programming \citep{Toussaint2015} optimize motions and manipulation decisions. More recently, MotionDisco \citep{taouil2026motiondisco} combines motion search with trajectory optimization, while OmniContact \citep{OmniContact} chains skills through contact-flow representations. We investigate whether these interactions can instead be discovered by steering the existing coordination of a frozen whole-body controller, without supplying a reference motion or training a new task policy.

\textbf{Reusing and steering pretrained behaviors.}
Latent skill methods such as SPiRL \citep{SPiRL}, OPAL \citep{OPAL}, SkiMo \citep{SkiMo}, {and ASE \citep{peng2022ase}} provide reusable primitives for downstream learning or planning. METRA \citep{METRA} and HILP \citep{HILP} learn behavioral spaces before downstream tasks are specified, while forward-backward representations \citep{FB} support reward-based policy inference without retraining. This direction underlies whole-body behavioral models including Meta Motivo \citep{tirinzoni2025zeroshot}, BFM-Zero \citep{li2026bfmzero}, and FB-MEBE \citep{FBMEBE}. {Beyond whole-body control, Play2Perfect \citep{lum2026play2perfect} pretrains a goal-conditioned dexterous hand policy through play and then fine-tunes it for each precise assembly task.} Their behavioral coverage provides a useful starting point, but finding commands for precise new interactions remains difficult. Adaptation methods address this through parameter updates, skill-based learning, or action corrections, as in OpenVLA \citep{OpenVLA}, SUPE \citep{SUPE}, and residual RL \citep{ResidualRL}. Closer to our setting, FlowDAgger \citep{murray2026flowdagger}, UniSteer \citep{UniSteer}, and Flow Reversal Steering \citep{FlowReversalSteering} map human or VLM corrections into the noise space of frozen generative policies. These works connect action-space guidance to latent execution. LJS instead discovers useful corrections from simulator-evaluated physical outcomes, without expert corrective actions or a learned steering policy.

\textbf{Searching through the policy's local inverse.}
CEM \citep{Rubinstein1999}, iCEM \citep{iCEM}, MPPI \citep{MPPI}, and DIAL-MPC \citep{DIALMPC} search through simulated outcomes; SkiMo \citep{SkiMo} and BFM-Zero \citep{li2026bfmzero} also demonstrate planning over learned behavioral spaces. Our question concerns how to construct useful proposals within a frozen controller. Structured exploration methods such as gSDE \citep{gSDE}, LATTICE \citep{LATTICE}, and DEP-RL \citep{DEPRL} organize perturbations for coordinated control, while Safe Mutations \citep{SafeMutations} and Policy Manifold Search \citep{PoMS} use output sensitivities to guide exploration. Drawing on local inverse mappings used in damped inverse kinematics \citep{Wampler1986} and operational-space control \citep{Khatib1987}, LJS couples joint-action proposals to latent execution through the policy Jacobian. CEM searches joint-command changes, a damped inverse maps them into latent perturbations, and simulator rollouts evaluate their physical effects. The Jacobian describes the frozen actor's local response, not contact dynamics. This coupling lets joint-action coordinates structure the search while the pretrained controller generates every executed command.

\section{Latent Jacobian Steer}
\label{sec:ljs}

\begin{figure}[t]
  \centering
  \includegraphics[width=0.8\linewidth]{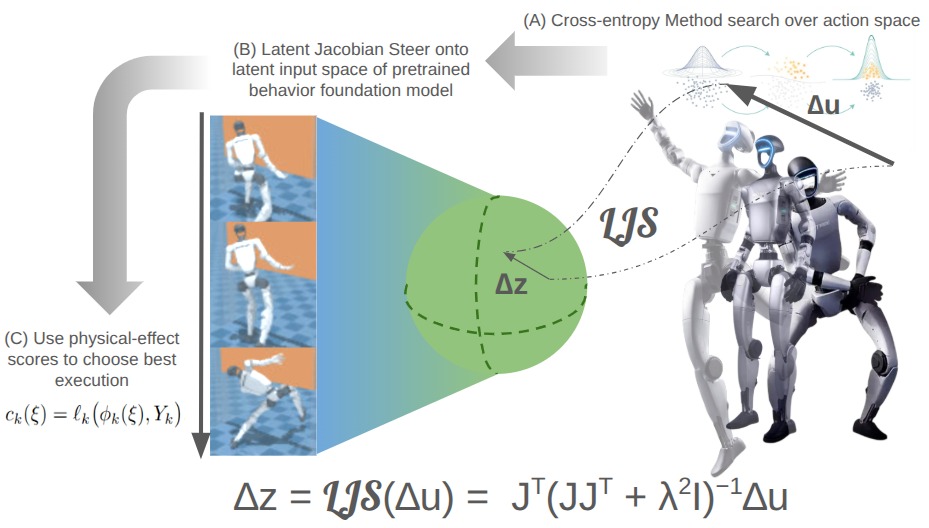}
  \caption{\textbf{Latent Jacobian Steer overview.}
  (A) CEM proposes joint-action perturbations $\Delta u$.
  (B) A damped inverse of the frozen policy Jacobian maps each proposal to a latent change $\Delta z$; the updated latent command is normalized and simulated through the pretrained controller.
  (C) Physical-effect scores rank the rollouts and guide CEM updates. The best admissible candidate is executed for a short window before replanning. The actor remains frozen throughout.}
  \label{fig:ljs_overview}
\end{figure}

{We propose} Latent Jacobian Steer (LJS){, which} couples {the} direct joint-action and latent search spaces. We search for task-relevant changes in {the} joint-action space, map them to changes in the latent command {via} the local Jacobian of the frozen policy, and evaluate the resulting behavior {using} physics simulation. Figure~\ref{fig:ljs_overview} gives an overview.

\textbf{Problem Formulation.} We consider humanoid control in a simulator as a finite-horizon Markov decision process $\mathcal{M}=(\mathcal{S},\mathcal{U},P,\rho_0,T)$, where $\mathcal{S}$ is the state space, $\mathcal{U}$ is the joint-command space, $P$ is the transition dynamics including the low-level controller, $\rho_0$ is the initial-state distribution, and $T$ is the task horizon. We specify the task objective separately via {a} \textit{physical-effect} sequence $\mathcal{T}$, defined below, whose scores guide the search and whose acceptance conditions determine task completion. The state $s_t$ includes the robot and environment configurations and velocities, together with the controller memory needed to make the transitions Markovian. For the 29-DOF humanoid considered here, $u_t\in\mathcal{U}\subseteq\mathbb{R}^{29}$ specifies joint-position targets, realized through a joint-space PD controller. We have a frozen {latent-conditioned} pretrained policy $u_t=\pi_\theta(h_t,z_t)$, where $h_t$ contains the robot's proprioceptive observations and action history, and $z_t\in\mathcal{Z}$ is its latent command. 
{For our particular purpose of demonstrating} the adaptation of the pretrained policy over interesting {out-of-distribution (OOD)} tasks, we begin by concretizing \textit{how to define a task itself}. It is worth noting that we chose this as our task definition for ease of expressing various tasks, and we are not proposing it as a canonical way to define loco-manipulation tasks. 
We define a task as an ordered sequence of physical effects $\mathcal{T}=(E_1,\ldots,E_K)$. Each effect $E_k$ specifies a physical measurement $\phi_k$, a desired target or target region $Y_k$, and a non-negative loss $\ell_k$. For a trajectory segment $\xi$, its physical-effect score is $c_k(\xi)=\ell_k\bigl(\phi_k(\xi),Y_k\bigr).$  The measurement may describe body or object displacement, orientation, contact forces, or conditions that must hold over the segment. When an effect requires several conditions simultaneously, we combine their losses: $c_k(\xi)=\sum_{j=1}^{m_k}w_{k,j} \ell_{k,j}\bigl(\phi_{k,j}(\xi),Y_{k,j}\bigr), w_{k,j}>0.$ We also specify an acceptance condition $A_k(\xi)$ for each effect. By default, an effect is accepted when all its component losses are zero; separate acceptance tolerances can be specified where needed. This distinguishes the score used to guide exploration from the condition used to decide whether the physical effect has been achieved. The task is satisfied when the executed trajectory admits ordered boundaries $0=t_0\leq t_1\leq\cdots\leq t_K\leq T$ such that
 $A_k\bigl(\xi_{t_{k-1}:t_k}\bigr)=1;k=1,\ldots,K.$
Any condition that must persist across multiple effects is included in each relevant effect. The {subsequent} search has access to privileged simulator state and contact information to evaluate the desired physical effects; the pretrained policy continues to receive its original observations. Our goal is to find a sequence of latent commands that satisfies these effects without changing $\theta${, that is, fine-tuning or retraining the policy}. Here, OOD tasks require contact interactions and physical effects that are not represented during the controller's pretraining.

The choice of pretrained controller is important for LJS. {Any} latent-conditioned policy or general motion tracker is not necessarily suitable simply because it supports diverse behaviors: its latent space must also be meaningfully steerable toward downstream objectives. For our main experiments, we use BFM-Zero \citep{li2026bfmzero}, which learns a shared latent space through unsupervised reinforcement learning with forward-backward representations and can be prompted using motions, goals, or rewards without task-specific retraining. This is particularly useful in our setting, where we specify the desired physical effect without providing a reference motion that realizes it. We therefore use LJS to explore how {to steer this latent space} toward new contact-rich behaviors.

\subsection{Our choice of latent-conditioned pretrained controller, BFM-Zero}
BFM-Zero \citep{li2026bfmzero} learns a family of whole-body behaviors through a latent-conditioned policy $\pi_\theta(h,z)$. Its forward-backward (FB) representation relates latent commands to the future states the policy visits, allowing motions, goals, and rewards to prompt the same controller without task-specific retraining. {We note that BFM-zero has \textit{not} been trained on environment interaction or loco-manipulation behaviors.}

For the {Unitree-G1-specific policy \citep{li2026bfmzero}} used {in our work}, the latent command lies on the sphere $\mathcal{Z}=\{z\in\mathbb{R}^{256}:\|z\|_2=16\}.$ The actor produces 29 joint-position targets at 50 Hz. LJS searches these latent commands while keeping the actor frozen. We defer the FB formulation and its prompting mechanisms to Appendix~\ref{app:bfm_background}.

\subsection{Defining the physical effects for a contact-rich OOD task}
{As described earlier,} we define a task through a small sequence of measurable physical outcomes. {To illustrate:} for shutter traversal, {this involves} placing the palms beneath the shutter, lifting it, moving the head and torso through the opening, {clearing it with the entire body, and standing upright on the far side}. Each effect specifies what must happen while leaving the movements that realize it to the search. For example, lifting can require a minimum shutter height and sufficient upward support, with a score penalizing shortfalls and a separate acceptance condition determining completion. {It is to be noted that defining these effects are not equivalent to dense reward engineering, as the underlying motion is not engineered and depends heavily on the frozen controller.}

We construct these effects stage-wise, using an LLM as a coding assistant to translate intuitive task descriptions into simulator measurements and scores. We inspect rollouts and refine definitions that admit unintended outcomes. We defer the mathematical score definitions, task-specific sequences, and acceptance conditions to Appendix~\ref{app:effect_definitions}.

\subsection{Coupling of the latent and direct action space}

Once the physical effect for a particular stage of a task is defined, our goal is to find latent commands that satisfy it. We do this by searching in joint-action space using CEM \citep{Rubinstein1999} and pulling the proposed changes back into the latent space through the frozen policy's local Jacobian.

\textbf{How is a joint-action change pulled back into the latent space?}
Let $u(h,z)$ denote the joint-position targets produced by the frozen policy, including its output scaling and latent normalization. At the current observation history $h$ and latent command $z$, we compute
\[
J(h,z)=\frac{\partial u(h,z)}{\partial z}
\in\mathbb{R}^{29\times256}.
\]
This is the policy Jacobian: it describes how the current joint targets change with the latent input while $h$ is held fixed. For a small latent change,
\[
u(h,z+\Delta z)-u(h,z)\approx J(h,z)\Delta z.
\]
Thus, given a proposed joint-action change $\Delta u$, we seek a latent change that approximately realizes it. Since $J$ is rectangular and may have weakly responsive directions, we use a damped least-squares inverse:
\[
\Delta z^\star
=
\underset{\Delta z}{\arg\min}\;
\|J\Delta z-\Delta u\|_2^2
+\lambda^2\|\Delta z\|_2^2,
\qquad \lambda>0.
\]
Setting the gradient to zero gives
\[
(J^\top J+\lambda^2I)\Delta z^\star=J^\top\Delta u,
\]
and hence
\[
\Delta z^\star
=
\underbrace{J^\top(JJ^\top+\lambda^2I)^{-1}}_{A}
\Delta u.
\]
The damping limits latent changes along directions in which the policy is insensitive. We then form the candidate
\[
z'=\Pi(z+A\Delta u),
\qquad
\Pi(v)=16\frac{v}{\|v\|_2}.
\]
The Jacobian includes differentiation through $\Pi$, so the local change accounts for the controller's spherical latent domain. This mapping approximates the policy's immediate output response.

\textbf{How do we search for joint-action changes?}
We adapt the CEM implementation in Hydrax \citep{kurtz2024hydrax} to search over joint-action perturbations. At each planning step, we initialize a zero-mean diagonal Gaussian and divide the candidate budget across several CEM rounds. Each sampled perturbation is mapped into a latent command: $\Delta u_i^{(q)}
\sim \mathcal{N}\!\left(\mu_q,\operatorname{diag}(s_q^2)\right), z_i^{(q)}=\Pi\!\left(z+A\Delta u_i^{(q)}\right).$
We simulate the candidates, rank them by their physical-effect scores, and refit the Gaussian to the best admissible joint-action proposals. Each round includes its current mean, so the first round also tests the unchanged command. A minimum standard deviation retains exploration; if no candidate is admissible, the distribution is left unchanged. We defer the full CEM updates and elite-selection rules to Appendix~\ref{app:cem_updates}.

\textbf{How do we score and execute a candidate?}
We simulate each candidate for $H$ control steps from the same saved simulator state and controller history, holding its latent command fixed while the frozen actor receives updated observations. We rank candidates by their mean physical-effect score over the latter half of the rollout, allowing time for the response to develop. Candidates violating full-segment constraints are rejected and excluded from the elite set.

After all CEM rounds, we execute the lowest-scoring admissible candidate for $L\leq H$ steps. Its latent command becomes the new anchor; we recompute the Jacobian and restart CEM with zero-mean perturbations. We advance only when the executed segment satisfies the current effect's acceptance conditions. Search ends when all effects are accepted, the budget is exhausted, or no admissible candidate is found. The exact rollout objective and search parameters are provided in the appendix.

\section{Experiments}
\label{sec:experiments}

We evaluate whether LJS can elicit contact-rich behaviors from a frozen whole-body controller, whether coupling joint-action proposals to latent execution improves task completion, and {whether the same holds for a different pretrained controller. We run two sets of experiments: three contact-rich whole-body tasks with the frozen BFM-Zero G1 humanoid controller in MuJoCo \citep{todorov2012mujoco}, and four precise assembly tasks with the frozen pretrained policy of Play2Perfect \citep{lum2026play2perfect}. Neither uses task-specific training or reference motions. Beyond task completion, we also examine search cost, motion quality, robustness to new starts and randomized physics, whether a search result can be used to train a closed-loop policy, a sanity check with minimal effect definitions, and search ablations.}

\subsection{Contact-rich Tasks}
\label{sec:tasks}

\begin{figure}[t]
  \centering
  \includegraphics[width=\linewidth]{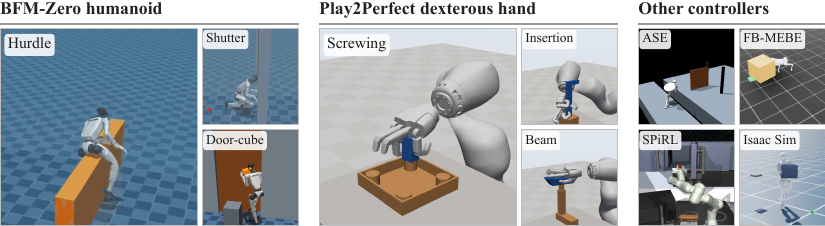}
  \caption{{\textbf{Behaviors elicited from frozen pretrained controllers with LJS.} Left: the BFM-Zero humanoid controller. Middle: the Play2Perfect dexterous hand policy. Right: further controllers and a second simulator (Appendix~\ref{app:other_controllers}). No controller was trained or fine-tuned for these tasks.}}
  \label{fig:task_overview}
\end{figure}

{\textbf{Humanoid tasks with BFM-Zero.}} We consider three tasks that require different combinations of contact establishment, whole-body support, object manipulation, and recovery {(Figure~\ref{fig:task_overview})}.

\textit{Hurdle traversal.} The humanoid must cross a rigid hurdle using its body for support. It establishes palm and leg contacts, moves into a supported straddling configuration, transfers its body across the obstacle, places both feet on the far-side floor, and returns to standing. The ten effects require alternating support configurations, leg clearance, and recovery after crossing.

\textit{Shutter opening and traversal.} The humanoid must place both palms beneath a vertically sliding shutter, lift it, move its head and torso through the opening, {clear the panel with its entire body, and stand upright on the far side}. These {five} effects combine precise palm placement and orientation with upward support{,} crouched whole-body movement{, and recovery}.

\textit{Door opening while carrying a cube.} The humanoid approaches a cube on a stand, grasps its opposing side faces, lifts it clear, and carries it to a closed door. It then uses the grasped cube to depress the handle and push the door open. The six effects require retaining the cube while manipulating another object.

{\textbf{Assembly tasks with Play2Perfect.}} {Our second set of experiments takes LJS beyond humanoids, to the pretrained play policy of Play2Perfect \citep{lum2026play2perfect}. This policy controls a Sharpa hand on a Franka arm through 29 joint targets and is conditioned on a 6-dimensional object goal. Play2Perfect fine-tunes the policy separately for each of its assembly tasks. We instead keep the pretrained policy frozen and only search over its goal input, on the same four tasks: placing the first and the second part of a beam assembly, screwing a leg onto its thread, and a tight peg insertion. Observe that the goal can be thought of loosely as a search variable that sequences and steers the policy over the whole episode, much like the latent command of BFM-Zero. Each task has three or four effects, for example picking up the leg, engaging the thread, turning it in, and releasing it for screwing. Here, LJS maps joint-action proposals through the Jacobian of the policy's joint targets with respect to its goal.}

{\textbf{Other controllers and simulators.}} {Since LJS only needs a frozen latent- or goal-conditioned policy and a simulator, we also show how it can be integrated with and used to steer the ASE \citep{peng2022ase} character controller to open a door and jump across a gap, the FB-MEBE \citep{FBMEBE} policy of a Go2 quadruped to push a box to a target, and the SPiRL \citep{SPiRL} skill decoder of a robot arm to take a cube out of a cabinet. Finally, we also run BFM-Zero with LJS in a second simulator, Isaac Sim \citep{mittal2025isaaclab}, to carry a breadbox (Figure~\ref{fig:task_overview}, Appendix~\ref{app:other_controllers}).}

\subsection{Defining the physical-effect scores}
\label{sec:defining_scores}

We construct effects stage-wise, beginning with an intuitive description of the required outcome. An LLM coding agent helps translate this description into simulator measurements, such as body-object distances, object orientation, contact forces, clearance, and velocity. These measurements define a score for ranking candidates and a separate acceptance condition for advancing to the next effect. They specify physical outcomes without prescribing the complete movement that realizes them.

We inspect the resulting rollouts {on development seeds} and revise definitions that admit unintended solutions. Every definition is human-reviewed and held fixed during evaluation; the LLM does not observe the search online or select commands. {The evaluation seeds are never used while designing the effects.} This back-and-forth required less than five hours to finalize a task's effects. All methods are evaluated with the same final task definitions. The mathematical definitions and an example Human--AI interaction card \citep{feng2026autonomousmathematicsresearch} are provided in Appendix~\ref{app:effect_definitions}; practical guidance is given in Appendix~\ref{app:tips}.

\subsection{Baselines and Evaluation}
\label{sec:baselines}

Our main comparisons keep CEM as the optimizer and change where proposals are made and how they are executed. We additionally ablate CEM while retaining the LJS coupling.

\textbf{CEM-LJS.} Our method searches over joint-action perturbations and maps each proposal into a latent command through the damped policy-Jacobian inverse. CEM fits its proposal distribution to the best simulated candidates, and the selected command is executed through the frozen actor.

\textbf{CEM direct-$z$.} We apply CEM directly to the 256-dimensional latent command, following the latent-optimization approach demonstrated by BFM-Zero \citep{li2026bfmzero}. This retains the optimizer and frozen-policy execution while removing the joint-action proposal space and Jacobian pull-back. {Note that, by itself, BFM-Zero's reward prompting cannot express our tasks, as its backward map sees only the robot's body, not the objects and contacts the effects measure (Appendix~\ref{app:bfm_background}).}

\textbf{CEM residual-BFM.} CEM searches for additive corrections to the frozen actor's joint-position targets. This retains joint-action proposals but removes the requirement that they be realized through latent commands, giving the search direct authority to override the actor's output.

\textbf{Mixed-Gaussian LJS.} This optimizer ablation retains the Jacobian pull-back and replaces CEM with Gaussian proposals. Below a physical-effect score threshold, finite-difference simulator probes estimate a targeted correction. The candidate population combines proposals around that correction and around zero, together with directional trials and the unchanged command.

{\textbf{CEM direct-goal.}} {For Play2Perfect, CEM searches the 6-dimensional goal with the same budgets.}

{\textbf{Evaluation.} We evaluate every method on ten seeds that were never used while defining the tasks. For the humanoid tasks, all methods share one search setting: each candidate is simulated for 40 control steps, the selected command is executed for 20 steps, each decision evaluates 1,024 candidates, and each effect allows at most 30 decisions. These settings are fixed across methods and seeds, to avoid any method-specific tuning. For Play2Perfect, candidates are simulated for 20 steps and executed for 10, with 1,024 candidates per decision. We report the mean number of accepted effects, the number of fully completed tasks, and the wall-clock search time on an NVIDIA RTX 4090.} Search cost counts simulated candidate control steps, including rejected candidates. For each effect $k$, we estimate the expected runtime (ERT) in these steps \citep{hansen2016cocoperformance}:
\[
\widehat{\mathrm{ERT}}_k
=\frac{\sum_{i=1}^{{10}} C_{i,k}}{\sum_{i=1}^{{10}}\mathbf{1}\{\text{run }i\text{ reaches effect }k\}}.
\]
Here, $C_{i,k}$ counts steps from task initialization to the first acceptance of effect $k$, or to termination if it is not reached. This estimates the cost of reaching the effect under independent restarts from the task's initial state, charging unsuccessful attempts as well.

\subsection{Results}
\label{sec:results}

{We ask the following questions to analyze whether LJS works as intended and what makes it robust.}

{\textbf{Does coupling the two spaces complete more of each task?}} {CEM-LJS completes the most effects on every humanoid task (Table~\ref{tab:main_results}), significantly so on the shutter and door-cube (Appendix~\ref{app:ten_seed}), and the gap grows with how precise the required contacts are. The shutter is the clearest example. Supporting and lifting the panel needs both hands placed beneath its edge and pushing upward, and CEM-LJS is the only method that gets through it. Direct latent search keeps the controller's coordination but cannot place the hands precisely enough to support the panel, so it almost never completes even the first effect. Residual search can move the hands to the right place, but overriding the joint targets also overrides the balance the controller provides. Its best run crawls under the panel with its torso on the floor and never stands back up (Figure~\ref{fig:shutter_qualitative}). On door-cube, CEM-LJS completes the task on every seed, while residual search never gets the cube to the door. The hurdle is the hardest task for every method, and CEM-LJS gets the furthest on average.}

\begin{table}[!ht]
  
  \centering
  \caption{\textbf{Humanoid task completion with BFM-Zero, ten seeds.} Mean accepted effects $\pm$ s.d.\ (fully completed tasks).}
  \label{tab:main_results}
  \small
  \setlength{\tabcolsep}{6pt}
  \renewcommand{\arraystretch}{1.12}
  \begin{tabular}{@{}lccc@{}}
    \toprule
    Method & Hurdle (10 effects) & Shutter (5 effects) & Door-cube (6 effects) \\
    \midrule
    \textbf{CEM-LJS} & \textbf{6.20 $\pm$ 3.43} (3/10) & \textbf{3.90 $\pm$ 2.08} (7/10) & \textbf{6.00 $\pm$ 0.00} (10/10) \\
    CEM direct-$z$ & 4.60 $\pm$ 2.99 (1/10) & 0.10 $\pm$ 0.32 (0/10) & 4.90 $\pm$ 1.45 (6/10) \\
    CEM residual-BFM & 5.30 $\pm$ 3.92 (3/10) & 1.50 $\pm$ 1.08 (0/10) & 1.70 $\pm$ 1.25 (0/10) \\
    \midrule
    Mixed-Gaussian LJS & 3.60 $\pm$ 3.53 (1/10) & 0.10 $\pm$ 0.32 (0/10) & 4.10 $\pm$ 2.47 (5/10) \\
    \bottomrule
  \end{tabular}
\end{table}

\begin{figure}[!ht]
\centering
\includegraphics[width=0.8\linewidth]{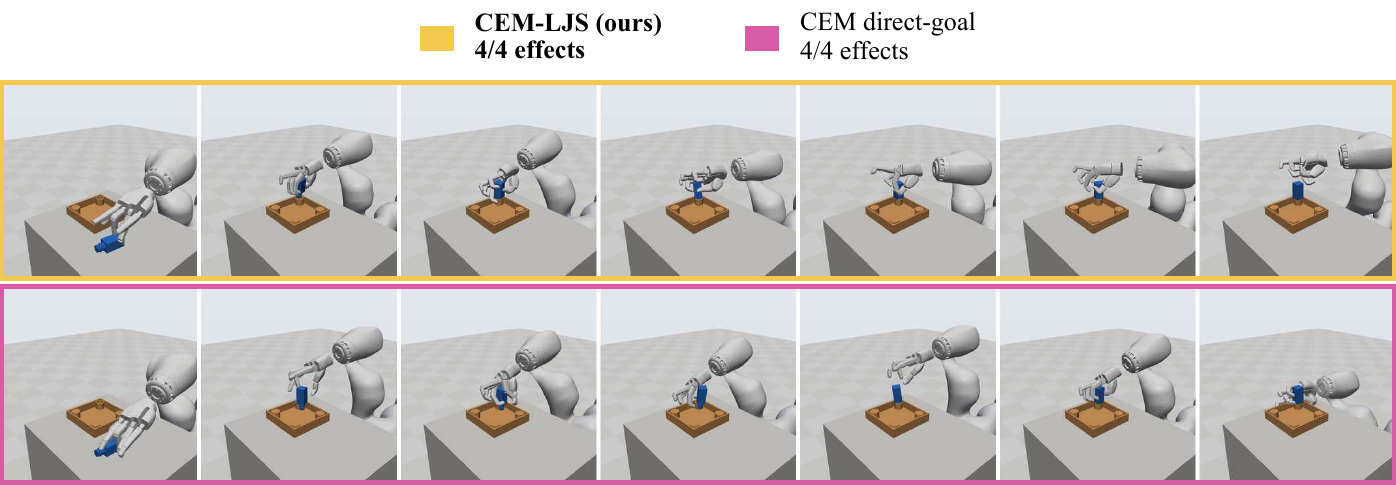}
\caption{\textbf{Screwing with the frozen Play2Perfect policy.} Seven evenly spaced frames from a successful run of each method. CEM-LJS completes screwing from 9 of 10 seeds and CEM direct-goal from 3. Appendix~\ref{app:play2perfect} shows the other assembly tasks.}
\label{fig:screwing_main}
\end{figure}

{\textbf{Does this carry over to another pretrained controller?}} {The same pattern holds for a very different controller and embodiment. Steering the frozen Play2Perfect policy, CEM-LJS completes 31 of the 40 assembly runs, against 21 for direct goal search (Table~\ref{tab:p2p_results}). It completes at least as many runs as direct goal search on every task, and far more on screwing (Figure~\ref{fig:screwing_main}). Where the two differ, CEM-LJS also reaches the later effects at a much lower search cost (Figure~\ref{fig:p2p_search_cost}). Searching the goal directly samples its six coordinates without regard to the hand, while LJS samples goal changes through their effect on the hand's joint targets, which matters most where the fingers must hold and turn the part.}

\begin{table}[!ht]
  
  \centering
  \caption{\textbf{Assembly task completion with the frozen Play2Perfect policy, ten seeds.} Same format as Table~\ref{tab:main_results}.}
  \label{tab:p2p_results}
  \small
  \setlength{\tabcolsep}{3pt}
  \renewcommand{\arraystretch}{1.12}
  \begin{tabular}{@{}lcccc@{}}
    \toprule
    Method & \stackhead{Beam,\\first part (3)} & \stackhead{Beam,\\second part (4)} & \stackhead{Screwing\\(4)} & \stackhead{Insertion\\(3)} \\
    \midrule
    \textbf{CEM-LJS} & \textbf{3.00 $\pm$ 0.00} (10/10) & \textbf{3.30 $\pm$ 1.25} (7/10) & \textbf{3.80 $\pm$ 0.63} (9/10) & \textbf{2.00 $\pm$ 1.05} (5/10) \\
    CEM direct-goal & \textbf{3.00 $\pm$ 0.00} (10/10) & 2.30 $\pm$ 1.49 (4/10) & 1.80 $\pm$ 1.62 (3/10) & 1.80 $\pm$ 1.03 (4/10) \\
    \bottomrule
  \end{tabular}
\end{table}

\begin{figure}[!ht]
  
  \centering
  \includegraphics[width=0.88\linewidth]{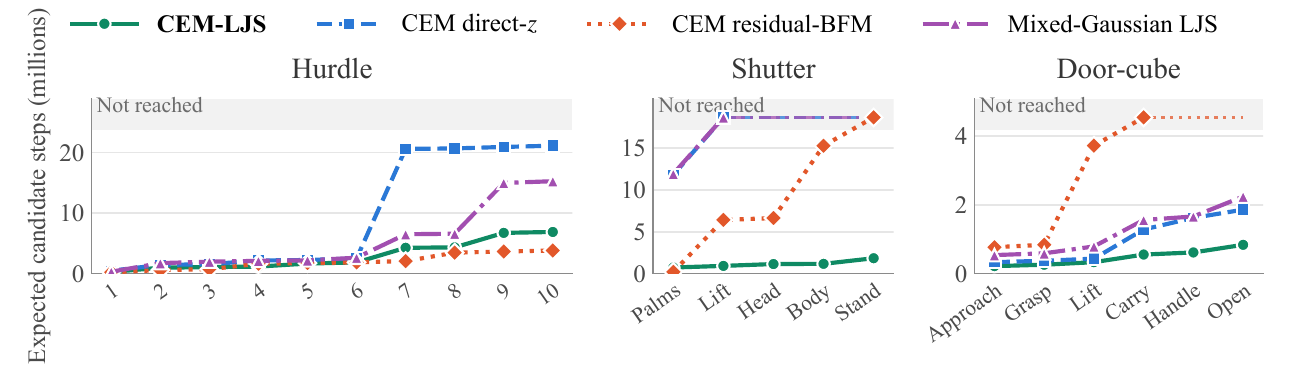}
  \caption{\textbf{Search cost to reach each physical effect, ten seeds.} Expected candidate control steps, including unsuccessful attempts (lower is better). Lines entering the ``not reached'' band mark effects that no run reached.}
  \label{fig:task_search_cost}
\end{figure}

\begin{figure}[!ht]
  
  \centering
  \includegraphics[width=0.88\linewidth]{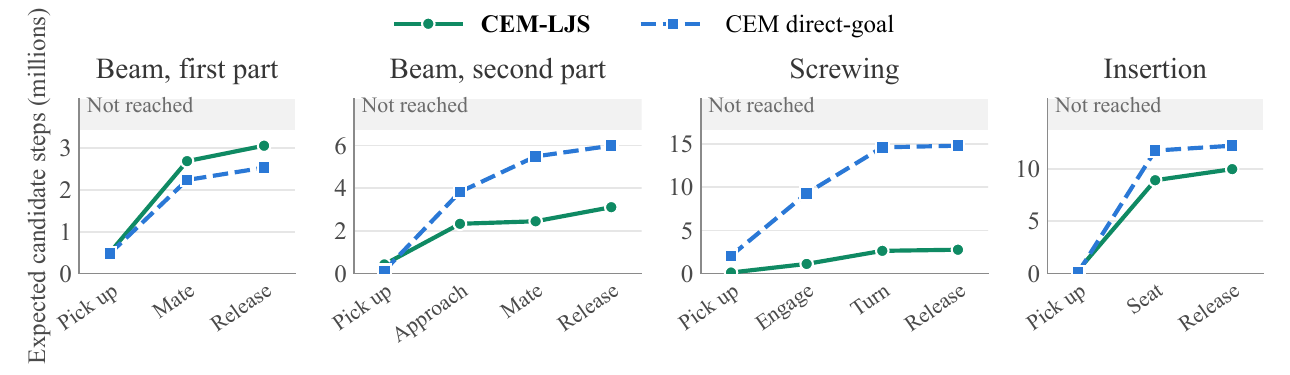}
  \caption{\textbf{Search cost to reach each effect with the frozen Play2Perfect policy, ten seeds.} Same measure as Figure~\ref{fig:task_search_cost}.}
  \label{fig:p2p_search_cost}
\end{figure}

{\textbf{How expensive is the search?}} {Figures~\ref{fig:task_search_cost} and~\ref{fig:p2p_search_cost} show where the search spends its effort. The differences appear at the contact-critical effects, such as lifting the shutter or lifting and carrying the cube, where CEM-LJS needs far fewer candidate steps than the baselines. On the hurdle, residual search reaches the later effects at a lower expected cost by correcting the joints directly, and this shows in its motion quality below. Most successful CEM-LJS searches finish within one to three minutes (Appendix~\ref{app:ten_seed}), which makes it practical to iterate on a task's effects.}

{\textbf{Is the resulting motion physically reasonable?} A search can still reach an effect with aggressive commands (Figure~\ref{fig:shutter_qualitative}). We therefore replay every run with the G1's joint and motor limits and compare CEM-LJS with residual search, which edits the joint targets directly and is the most likely to strain them (Table~\ref{tab:motion_main}). Comparing the effects that both methods complete, we find that CEM-LJS uses about 40\% less mechanical power, has less than half the joint jerk, spends far less time at the motor torque limits, and slips much less under load, and no run of any method exceeds the G1's joint velocity limits. To judge how natural the motion is, we also score it with BFM-Zero's own motion discriminator, which rates CEM-LJS as more human-like even though neither search optimizes for it. This is fairly intuitive, as every LJS proposal is executed through the frozen controller and keeps its coordination. Further, since a task is just a set of physical-effect scores, we can also shape the motion at its roots by adding the discriminator as a cost to the hurdle's scores, which gives a calmer and more human-like crossing with about 30\% less mechanical power (Appendix~\ref{app:motion_quality}).}

\begin{table}[!ht]
\centering
\small
\caption{\textbf{Motion quality of CEM-LJS and CEM residual-BFM} over the 65 effects both complete (medians), with the G1's joint and motor limits. Appendix~\ref{app:motion_quality} gives all 13 measures and the tests.}
\label{tab:motion_main}
\begin{tabular}{@{}lccc@{}}
\toprule
Measure (per effect) & CEM-LJS & CEM residual-BFM & CEM-LJS better on \\
\midrule
Mechanical power (W/kg) & \textbf{1.59} & 2.61 & 56/65 \\
Joint jerk, RMS (rad/s$^3$) & \textbf{782} & 1660 & 54/65 \\
Time at the motor torque limit & \textbf{0.06} & 0.20 & 43/65 \\
Foot slip under load (m/s) & \textbf{0.020} & 0.112 & 55/65 \\
Human-likeness (higher is better) & \textbf{$-$0.89} & $-$2.40 & 57/65 \\
\bottomrule
\end{tabular}
\end{table}

{\textbf{Do the same definitions hold up under changes?}} {We observe that the effect definitions carry over to new starting configurations without any retuning. When the shutter task starts from moved and turned positions, CEM-LJS still gets through from 8 of 10 starts, while no baseline gets through more than twice (Appendix~\ref{app:robustness}). For all practical purposes, a single search can settle in a poor local optimum, so we retry a failed effect, from an earlier state if needed, and score every candidate under randomized friction, mass, motor strength, and control delay. CEM-LJS then completes all 30 runs across the three tasks, while the best baseline completes 22 (Table~\ref{tab:retries} and Appendix~\ref{app:robustness}). This largely removes the factor of luck.}

\begin{table}[!ht]

\centering
\small
\caption{\textbf{Retries with randomized physics, ten seeds per task.} Fully completed tasks and CEM-LJS search time.}
\label{tab:retries}
\begin{tabular}{@{}lccccc@{}}
\toprule
Task & CEM-LJS & \stackhead{CEM\\direct-$z$} & \stackhead{CEM\\residual-BFM} & \stackhead{Mixed-\\Gaussian LJS} & CEM-LJS time \\
\midrule
Hurdle (10) & \textbf{10/10} & 6/10 & 7/10 & 2/10 & 479\,s (221--856\,s) \\
Shutter (5) & \textbf{10/10} & 6/10 & 0/10 & 3/10 & 328\,s (137--887\,s) \\
Door-cube (6) & \textbf{10/10} & \textbf{10/10} & 2/10 & \textbf{10/10} & 145\,s (128--161\,s) \\
\midrule
All & \textbf{30/30} & 22/30 & 9/30 & 15/30 & \\
\bottomrule
\end{tabular}
\end{table}

\begin{figure}[!ht]
  \centering
  \includegraphics[width=\linewidth]{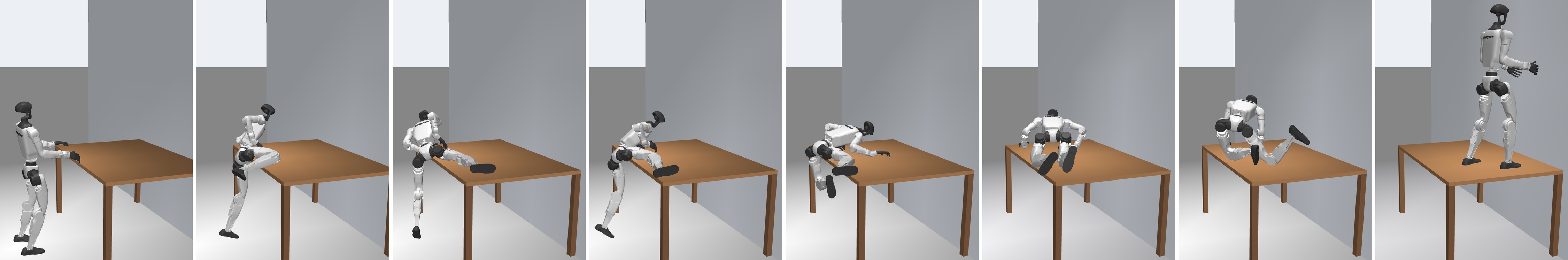}
  \caption{\textbf{A closed-loop policy trained from one CEM-LJS rollout} climbs onto a 0.75\,m table and stands up. It runs here in standard MuJoCo, a simulator it was not trained in, on a randomized table.}
  \label{fig:table_climb}
\end{figure}

\textbf{Can what LJS finds be used downstream?} The searches above hold up under randomized physics, but they rely on privileged simulator access, so the latent-command sequences they produce are not controllers by themselves. We therefore ask whether a single search result is enough to train one. In a new task, the G1 climbs onto a 0.75\,m table pushed against a wall and stands up on it (Figure~\ref{fig:table_climb}). CEM-LJS finds one climb under randomized physics, and we train a closed-loop policy to track it with SuperMimic \citep{liao2026supermimic}. Using only its IMU, its joint encoders and the table edge seen from its head camera, the policy climbs and stands in 96\% of trials with the table, the robot's dynamics and its sensor readings randomized. It also succeeds in every trial in standard MuJoCo, a simulator it was not trained in. Since the policy only has to follow a climb that already works under varied physics, LJS outputs can directly serve as demonstrations for closed-loop controllers (Appendix~\ref{app:closed_loop}).

\begin{figure}[!ht]
  \centering
  \includegraphics[width=0.46\linewidth,trim=448 62 40 55,clip]{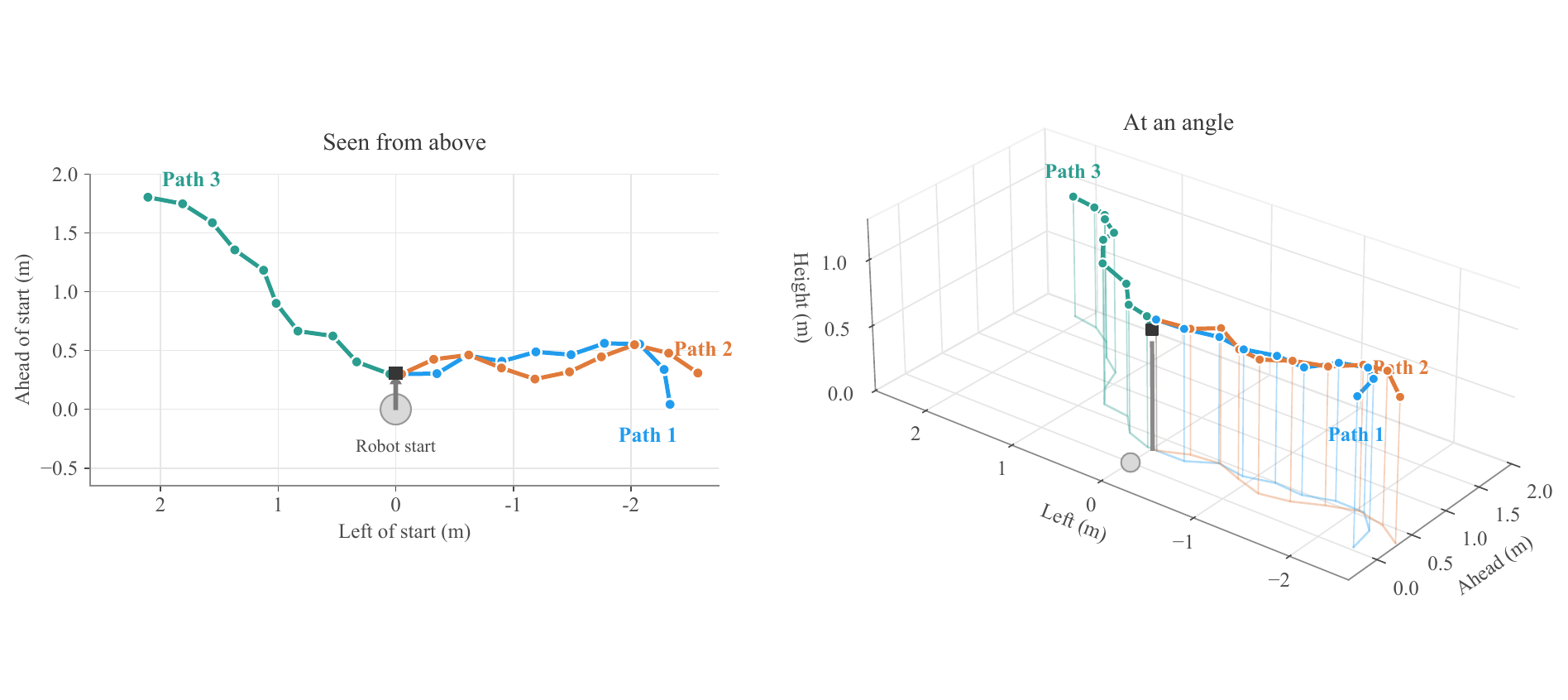}
  \caption{\textbf{The three generated paths} for the sanity check.}
  \label{fig:carry_paths}
\end{figure}

{\textbf{Does the comparison depend on how the effects are written?}} {As a sanity check, we define a task whose definition leaves nothing to tune. The humanoid carries a 3\,kg box, held between its palms, along a path of points in 3D space. Each effect only asks the box to reach the next point, and the box must never touch the floor. This minimal definition imposes nothing on the robot or on how it handles the box, and it is the same for every method. With the same retries for every method, CEM-LJS completes the large majority of runs, while no baseline completes more than a fifth of them (Table~\ref{tab:sanity_main} and Figure~\ref{fig:carry_paths}).}

\begin{table}[!ht]
  \centering
  \caption{\textbf{Sanity check with a minimal definition}, three paths with five seeds each.}
  \label{tab:sanity_main}
  \small
  \renewcommand{\arraystretch}{1.12}
  \begin{tabular}{@{}lcc@{}}
    \toprule
    Method & Runs completed & Mean points reached (of 10) \\
    \midrule
    \textbf{CEM-LJS} & \textbf{12/15} & \textbf{9.1} \\
    CEM direct-$z$ & 2/15 & 2.5 \\
    CEM residual-BFM & 3/15 & 4.8 \\
    \midrule
    Mixed-Gaussian LJS & 1/15 & 1.5 \\
    \bottomrule
  \end{tabular}
\end{table}

{\textbf{Ablations.}} {To check that the benefit comes from the coupling itself, we vary the optimizer and the proposal space on all three humanoid tasks with ten seeds each (Table~\ref{tab:why_coupling}). With the same optimizer, proposals pulled back through the Jacobian complete more effects than proposals in the latent space in 11 of the 12 task and optimizer pairs (Appendix~\ref{app:ablations}). A random latent subspace of the same dimension as the LJS proposals completes no task. We then ask which part of the coupling matters. Rotating the joint-action axes of CEM-LJS changes little, but searching the Jacobian's own latent directions without the rescaling of LJS completes less than half as many tasks, in line with the analysis in Appendix~\ref{app:why_ljs}. Finally, to make sure that the baselines are not simply mistuned, we also vary their proposal scales on the shutter, and none of them gets through at any scale (Appendix~\ref{app:settings_sweeps}).}

\begin{table}[!ht]

\centering
\small
\caption{\textbf{What makes the coupling work}, ten seeds per task: fully completed tasks and mean accepted effects.}
\label{tab:why_coupling}
\begin{tabular}{@{}lccc@{}}
\toprule
Search & Shutter (5) & Hurdle (10) & Door-cube (6) \\
\midrule
\textbf{CEM-LJS} & \textbf{7/10}, 3.9 & \textbf{3/10}, 6.2 & \textbf{10/10}, 6.0 \\
CEM-LJS, rotated proposal axes & 7/10, 3.6 & 0/10, 5.7 & 10/10, 6.0 \\
CEM over the Jacobian's latent directions & 3/10, 1.6 & 0/10, 4.1 & 6/10, 4.8 \\
\midrule
CEM direct-$z$ & 0/10, 0.1 & 1/10, 4.6 & 6/10, 4.9 \\
Random 29-D latent subspace & 0/10, 0.0 & 0/10, 0.8 & 0/10, 1.9 \\
\bottomrule
\end{tabular}
\end{table}

\FloatBarrier
\section{Conclusion}

  We presented Latent Jacobian Steer (LJS), a method for eliciting contact-rich OOD behaviors from frozen latent-conditioned humanoid controllers. LJS proposes physically meaningful changes in joint-action space, pulls them back into the latent space through a damped local inverse of the policy Jacobian, and evaluates the resulting latent commands through simulation. LJS can therefore be interpreted as an interface that makes behavioral foundation models \textit{precise}. Importantly, LJS produces a sequence of latent commands $(z_1,\ldots,z_M)$ rather than an open-loop trajectory of joint actions. While each command is held fixed for a short execution window, the frozen policy continues to receive updated observations and generate new joint targets. These latent sequences can also serve as contact-rich demonstrations for behavioral cloning, reinforcement learning, or policy distillation; initialize subsequent task-specific optimization; and help probe which behaviors are present in a pretrained controller but difficult to access through direct prompting.

  The current method is an offline behavior-generation procedure. It requires privileged simulator measurements, saved-state resets, and many parallel candidate rollouts, making it unsuitable for direct real-time deployment. Its success is also limited by the behavioral support of the frozen controller, while the decomposition into physical effects remains human-specified and can admit unintended solutions. We show that a single LJS-generated behavior can already be distilled into a closed-loop policy that carries over to a second simulator, and a natural next step is the sim-to-real evaluation of such policies. We hope this simple and elegant method can give rise to new data-generation pipelines as well as learning paradigms that make use of pretrained controllers while not losing the authority of exploring in direct action space. 

\subsection*{AI use statement}

We used generative AI tools to polish manuscript language, assist with literature discovery and summarization, and write and debug research code. AI assistance also supported the translation of author-provided natural-language task specifications into physical-effect measurements, scoring functions, and acceptance conditions, including iterative revisions informed by author feedback and analysis of simulation outputs. The research ideas, problem formulation, and overall research direction originated with the authors.

We reviewed all AI-assisted work, including manuscript text, code, physical-effect definitions, and interpretations of experimental results. The authors retained responsibility for methodological decisions and the claims reported in the paper. We take full responsibility for the final content of this work, including text, claims, and artifacts produced with the assistance of generative AI.

\subsection*{Reproducibility statement}

Sections~\ref{sec:ljs} and~\ref{sec:experiments} describe the method, tasks, baselines, and evaluation protocol. Appendix~\ref{app:cem_updates} specifies the CEM updates, Appendix~\ref{app:effect_definitions} gives the physical-effect scores and acceptance conditions, and Appendix~\ref{app:ten_seed} lists the seeds, search settings, and the outcome and search time of every evaluation run. Appendix~\ref{app:closed_loop} describes the closed-loop policy and its evaluation. All methods use the same finalized task definitions, with the frozen BFM-Zero G1 controller for the humanoid tasks and the frozen Play2Perfect policy for the assembly tasks. The evaluations ran one search at a time on an NVIDIA RTX 4090. Videos of all evaluation runs and the additional results are on the project page, \url{https://latent-jacobian-steer.github.io/}, where the code will also be released.

\subsection*{Acknowledgments}

This research was partly supported by NUS under their Robotics Grand Challenge, as well as by ST Engineering. S K Swaminathan was supported by the International Relations Office of IIT Kharagpur through its semester-away program. We thank Jeric Lew for insightful discussions about the method, and Prof.\ Aritra Hazra of IIT Kharagpur for supporting and guiding S K Swaminathan throughout the semester-away project, and for his insights.

\bibliography{iclr2027_conference}
\bibliographystyle{iclr2027_conference}

\appendix
\FloatBarrier
\section{Theory}

\subsection{Forward-backward representations and BFM-Zero}
\label{app:bfm_background}

We briefly describe the forward-backward (FB) representation underlying BFM-Zero \citep{li2026bfmzero}. For simplicity, we write the expressions in terms of the state $s$, suppressing the distinction between privileged state and observation history.

The FB representation learns a forward mapping $F$ and a backward mapping $B$ that together approximate the discounted future-state occupancy under the latent-conditioned policy $\pi_z$:
\[
M^{\pi_z}(s,u,\mathrm{d}s')
\approx F(s,u,z)^\top B(s')\,\rho(\mathrm{d}s'),
\]
where $\rho$ is the reference state distribution and
\[
M^{\pi_z}(s,u,X)
=
\mathbb{E}\!\left[
\sum_{t=0}^{\infty}\gamma^t
\mathbf{1}\{s_{t+1}\in X\}
\,\middle|\,
s_0=s,\ u_0=u,\ \pi_z
\right].
\]
Thus, $F$ and $B$ describe which future states a latent-conditioned policy is expected to visit.

For a state-based reward $r$, using the convention that rewards are evaluated at the next state, this factorization gives
\[
Q_r^{\pi_z}(s,u)
\approx
F(s,u,z)^\top
\mathbb{E}_{s'\sim\rho}\!\left[B(s')r(s')\right].
\]
The policy is trained to favor actions with large $F(s,u,z)^\top z$. Consequently, the reward-weighted backward features provide a latent command for a downstream reward:
\[
z_r=\mathbb{E}_{s\sim\rho}\!\left[B(s)r(s)\right].
\]
This provides a way to prompt the pretrained policy without task-specific retraining. Similarly, a desired state can be embedded through $B$, while a reference motion can be represented by a sequence of state embeddings. The resulting commands are normalized to the controller's latent domain before execution.

These prompting mechanisms explain the behavioral structure of the pretrained controller. LJS instead searches latent commands through the frozen actor's Jacobian and simulator-evaluated physical effects; it does not use $F$ or $B$ during search. {Reward prompting also cannot express our tasks. The backward map $B$ of the G1 controller takes only the robot's own state, namely its joint positions and velocities and the positions, orientations, and velocities of its bodies, so a reward on the shutter, the cube, the door, or contact forces is not a function of its input. A prompted latent is also a single command for the whole task, while our tasks are ordered sequences of effects.}

\subsection{CEM updates for joint-action search}
\label{app:cem_updates}

We adapt the CEM implementation in Hydrax \citep{kurtz2024hydrax} to search over joint-action perturbations. At each planning step, we initialize
\[
\mu_0=0,
\qquad
s_0=\sigma\mathbf{1}_{29},
\]
where $\sigma>0$ is the initial exploration scale. We divide the candidate budget across $R$ rounds. In round $q$, we draw $N_q$ proposals and map them into latent commands:
\[
\Delta u_i^{(q)}
\sim
\mathcal{N}\!\left(\mu_q,\operatorname{diag}(s_q^2)\right),
\qquad
z_i^{(q)}=\Pi\!\left(z+A\Delta u_i^{(q)}\right).
\]
One proposal is set to the current mean,
$\Delta u_1^{(q)}=\mu_q$, so the first round includes the unchanged command. The latent anchor $z$ and pull-back map $A$ remain fixed throughout these rounds.

We simulate each candidate and rank admissible proposals by their physical-effect scores. Let $\mathcal{V}_q$ denote the admissible candidates. The elite set $\mathcal{E}_q$ contains the lowest-scoring
\[
n_q^{\mathrm{elite}}
=
\min\!\left(
|\mathcal{V}_q|,
\max(2,\lceil\kappa N_q\rceil)
\right)
\]
proposals, where $\kappa\in(0,1]$ is the elite fraction. When the elite set is nonempty, we update
\[
\mu_{q+1}
=
\frac{1}{|\mathcal{E}_q|}
\sum_{i\in\mathcal{E}_q}\Delta u_i^{(q)},
\]
\[
s_{q+1,j}
=
\max\!\left(
s_{\min},
\sqrt{
\frac{1}{|\mathcal{E}_q|}
\sum_{i\in\mathcal{E}_q}
\left(\Delta u_{i,j}^{(q)}-\mu_{q+1,j}\right)^2
}
\right),
\qquad j=1,\ldots,29.
\]
The floor $s_{\min}>0$ prevents the proposal distribution from collapsing completely. If no candidate is admissible, we retain the previous distribution:
\[
\mu_{q+1}=\mu_q,
\qquad
s_{q+1}=s_q.
\]

After all rounds, we select the lowest-scoring admissible candidate evaluated across the entire planning step. Its latent command is executed for a short window before recomputing the Jacobian and restarting CEM from the resulting state. If no admissible candidate was found, the search terminates without declaring task completion.

\FloatBarrier
\section{Why does LJS work, mathematically?}\label{app:why_ljs}

A small change in the latent command can move some joints strongly
and others barely at all. LJS accounts for this unequal sensitivity
when constructing its proposals. Intuitively, if we choose some random (including 29-D) space over the action space itself, some degrees of freedom might be unfairly disadvantaged. This is why we argue that LJS gives exploration enough authority to discover precise changes.

At the current observation $s$ and latent command $z$, let
\[
    J=\frac{\partial \pi(s,z)}{\partial z}.
\]
Locally, a latent change $\Delta z$ produces the action change
$\Delta u \approx J\Delta z$. LJS samples a desired action change
$\delta u$ and solves
\[
    \min_{\Delta z}
    \|J\Delta z-\delta u\|^2+\lambda^2\|\Delta z\|^2.
\]
Its solution is
\[
    \Delta z=J^\top(JJ^\top+\lambda^2I)^{-1}\delta u.
\]
Thus, LJS tries to realize the proposed action change without
requiring an excessively large latent change.

Why is this different from searching in another 29-dimensional
space? Consider even the latent directions selected by the Jacobian
itself. Write $J=U\Sigma V^\top$, with nonzero singular values $s_i$.
Sampling $\Delta z=V\epsilon$, where
$\epsilon\sim\mathcal N(0,\sigma^2I)$, produces
\[
    J\Delta z=U\Sigma\epsilon.
\]
The action variance along direction $i$ is $\sigma^2s_i^2$.
Consequently, some action directions receive much less exploration
than others, despite equal sampling variance in the search space.

With LJS and $\delta u\sim\mathcal N(0,\sigma^2I)$, that variance is
instead
\[
    \sigma^2
    \left(\frac{s_i^2}{s_i^2+\lambda^2}\right)^2.
\]
For $s_i\gg\lambda$, it is approximately $\sigma^2$: LJS compensates
for the controller's unequal sensitivities. For $s_i\ll\lambda$,
damping suppresses the correction rather than demanding a large
latent displacement. These action directions can be combinations
of joints, not necessarily individual joints.

The benefit is therefore more than reducing the search dimension.
LJS makes comparable proposals produce approximately comparable
immediate action changes in directions the controller can realize.
An arbitrary low-dimensional latent space need not do this. Appendix~\ref{app:why_coupling} tests this experimentally.

\subsection{What makes the coupling work}
\label{app:why_coupling}

{Appendix~\ref{app:why_ljs} explains why LJS works. The policy responds very strongly to some latent directions and barely to others, so equally sized random changes in the latent command mostly move the joints along a few directions. The damped inverse of LJS corrects for this, so that equally sized joint-action proposals produce comparably sized changes in the joints. Two simpler explanations could also account for the results of Table~\ref{tab:ablations_ten}, and we test both on all three tasks with the protocol of Appendix~\ref{app:ten_seed}. First, CEM adapts its search width separately along each axis it samples in, so LJS might help simply because the joint axes happen to be convenient axes for CEM. To test this, we rotate the joint-action proposals of CEM-LJS by one fixed random rotation. This keeps the initial proposals the same and only changes the axes along which CEM adapts. Second, LJS might help only because it searches the 29 latent directions that the policy actually responds to. To test this, we let CEM search along exactly these directions, the right-singular vectors of the policy Jacobian ($\Delta z=Vc$), but without the rescaling of LJS.}

{With rotated axes, CEM-LJS completes as many shutter and door-cube tasks as before and a similar number of hurdle effects, although none of its hurdle runs reaches the final effects (Table~\ref{tab:why_coupling}). Over the three tasks, it completes 17 of 30 runs against 20 for CEM-LJS (Fisher's exact test, $p=0.60$). Searching along the Jacobian's latent directions without the rescaling behaves very differently. It completes only 9 of 30 runs ($p=0.009$) and fewer effects on every task, though still far more than a random 29-dimensional latent subspace, which completes none. A natural suspicion is that its proposals are simply too small. Scaling them up until they move the joints as much as CEM-LJS proposals, however, still leaves it at 3 of 10 shutters. Together, these tests point to the two parts of LJS described above: the Jacobian selects the latent directions the policy responds to, and the damped inverse balances how strongly each of them moves the joints.}

\FloatBarrier
\section{Evaluation with ten seeds}
\label{app:ten_seed}

{\textbf{Seeds and settings.} The humanoid results use the seeds 12, 10, 19, 2027, 2004, 1264, 2681, 752, 2360, and 2981. None of them was used while defining the physical effects. All four methods use the same effects and the same search budget on every task: candidates are simulated for 40 control steps, the selected command is executed for 20 steps, each decision evaluates 1,024 candidates (1,023 sampled and the unchanged command) over four CEM rounds, and each effect allows at most 30 decisions. Because the selected candidate has already been simulated, we continue from its simulated state at the end of the execution window instead of simulating it again. When a CEM round contains no candidate that satisfies the hard constraints of the current effect, the losses of those constraints guide the sampling of the next round, and a candidate is only executed if it satisfies every hard constraint. Each method uses the same proposal scales on every seed. Every search runs alone on an NVIDIA RTX 4090 in a fresh process.}

{\textbf{Search time.} Table~\ref{tab:ten_seed_times} gives the wall-clock search time of the successful runs. The search time and outcome of every run are listed in Table~\ref{tab:seed_results_ten}, and the videos of all evaluation runs are available on the project page.}

\begin{table}[h]

\centering
\small
\caption{\textbf{Wall-clock search time of successful runs, ten seeds.} Mean (fastest), on an NVIDIA RTX 4090 with one search at a time. A dash marks a method that never completes the task.}
\label{tab:ten_seed_times}
\begin{tabular}{@{}lccc@{}}
\toprule
Method & Hurdle & Shutter & Door-cube \\
\midrule
CEM-LJS & 137\,s (104\,s) & 77\,s (54\,s) & 61\,s (55\,s) \\
CEM direct-$z$ & 152\,s (152\,s) & -- & 72\,s (58\,s) \\
CEM residual-BFM & 89\,s (80\,s) & -- & -- \\
Mixed-Gaussian LJS & 189\,s (189\,s) & -- & 82\,s (62\,s) \\
\bottomrule
\end{tabular}
\end{table}

{\textbf{Statistical comparison.} To check that the differences in Table~\ref{tab:main_results} are not seed-to-seed noise, we compare the accepted effects of CEM-LJS with each baseline over the ten seeds with a two-sided Mann--Whitney test (Table~\ref{tab:ten_seed_tests}). On the shutter and door-cube, CEM-LJS differs from every baseline. On the hurdle, the methods overlap more, and CEM-LJS has the highest mean.}

\begin{table}[h]

\centering
\small
\caption{\textbf{CEM-LJS against each baseline, ten seeds.} Two-sided Mann--Whitney $p$-values on the accepted effects per seed.}
\label{tab:ten_seed_tests}
\begin{tabular}{@{}lccc@{}}
\toprule
CEM-LJS vs. & Hurdle & Shutter & Door-cube \\
\midrule
CEM direct-$z$ & 0.29 & \textbf{0.001} & \textbf{0.034} \\
CEM residual-BFM & 0.70 & \textbf{0.020} & \textbf{$<$0.001} \\
Mixed-Gaussian LJS & 0.12 & \textbf{0.001} & \textbf{0.015} \\
\bottomrule
\end{tabular}
\end{table}

\begin{table}[h]

\centering
\scriptsize
\setlength{\tabcolsep}{3pt}
\caption{\textbf{Wall-clock search time and outcome of every BFM-Zero evaluation run, ten seeds.} Each entry gives the accepted effects / task total and, in parentheses, the search time in seconds on an NVIDIA RTX 4090. Fully completed tasks are in bold.}
\label{tab:seed_results_ten}
\resizebox{\linewidth}{!}{%
\begin{tabular}{@{}llcccccccccc@{}}
\toprule
Task & Method & 12 & 10 & 19 & 2027 & 2004 & 1264 & 2681 & 752 & 2360 & 2981 \\
\midrule
Hurdle & CEM-LJS & 1/10\,(88) & 8/10\,(162) & 6/10\,(157) & \textbf{10/10\,(104)} & 4/10\,(104) & \textbf{10/10\,(110)} & 1/10\,(88) & \textbf{10/10\,(198)} & 6/10\,(78) & 6/10\,(119) \\
 & CEM direct-$z$ & 6/10\,(175) & 1/10\,(45) & \textbf{10/10\,(152)} & 6/10\,(98) & 6/10\,(136) & 3/10\,(113) & 6/10\,(142) & 1/10\,(108) & 6/10\,(175) & 1/10\,(35) \\
 & CEM residual-BFM & \textbf{10/10\,(93)} & 1/10\,(34) & \textbf{10/10\,(94)} & 7/10\,(88) & 3/10\,(115) & 3/10\,(124) & 1/10\,(31) & 7/10\,(80) & \textbf{10/10\,(80)} & 1/10\,(20) \\
 & Mixed-Gaussian LJS & 1/10\,(43) & 6/10\,(130) & 1/10\,(51) & 1/10\,(46) & 1/10\,(49) & 6/10\,(152) & 1/10\,(48) & 8/10\,(244) & \textbf{10/10\,(189)} & 1/10\,(58) \\
\midrule
Shutter & CEM-LJS & 4/5\,(103) & \textbf{5/5\,(86)} & \textbf{5/5\,(105)} & \textbf{5/5\,(61)} & 0/5\,(86) & \textbf{5/5\,(75)} & \textbf{5/5\,(75)} & 0/5\,(84) & \textbf{5/5\,(54)} & \textbf{5/5\,(83)} \\
 & CEM direct-$z$ & 1/5\,(122) & 0/5\,(79) & 0/5\,(80) & 0/5\,(83) & 0/5\,(80) & 0/5\,(82) & 0/5\,(83) & 0/5\,(80) & 0/5\,(78) & 0/5\,(80) \\
 & CEM residual-BFM & 1/5\,(91) & 1/5\,(103) & 1/5\,(97) & 1/5\,(98) & 1/5\,(97) & 4/5\,(198) & 1/5\,(96) & 1/5\,(91) & 3/5\,(156) & 1/5\,(94) \\
 & Mixed-Gaussian LJS & 0/5\,(77) & 0/5\,(76) & 0/5\,(75) & 0/5\,(77) & 0/5\,(79) & 0/5\,(73) & 0/5\,(80) & 0/5\,(76) & 1/5\,(106) & 0/5\,(72) \\
\midrule
Door-cube & CEM-LJS & \textbf{6/6\,(60)} & \textbf{6/6\,(55)} & \textbf{6/6\,(57)} & \textbf{6/6\,(58)} & \textbf{6/6\,(55)} & \textbf{6/6\,(56)} & \textbf{6/6\,(71)} & \textbf{6/6\,(55)} & \textbf{6/6\,(70)} & \textbf{6/6\,(69)} \\
 & CEM direct-$z$ & 3/6\,(107) & \textbf{6/6\,(69)} & 3/6\,(49) & \textbf{6/6\,(78)} & 4/6\,(81) & 3/6\,(112) & \textbf{6/6\,(66)} & \textbf{6/6\,(79)} & \textbf{6/6\,(58)} & \textbf{6/6\,(82)} \\
 & CEM residual-BFM & 2/6\,(109) & 3/6\,(33) & 2/6\,(92) & 0/6\,(87) & 3/6\,(35) & 0/6\,(85) & 2/6\,(96) & 3/6\,(34) & 2/6\,(104) & 0/6\,(80) \\
 & Mixed-Gaussian LJS & 3/6\,(39) & \textbf{6/6\,(67)} & 0/6\,(75) & 5/6\,(84) & 3/6\,(107) & \textbf{6/6\,(82)} & \textbf{6/6\,(62)} & \textbf{6/6\,(86)} & \textbf{6/6\,(114)} & 0/6\,(81) \\
\bottomrule
\end{tabular}}
\end{table}

\FloatBarrier
\section{Search ablations}
\label{app:ablations}

{\textbf{Ten seeds on all three tasks.} We run every search variant under the protocol of Appendix~\ref{app:ten_seed}, with the same seeds, budgets, and effect definitions. Only the proposal coordinates, the population update, or the mix of proposals change between variants (Table~\ref{tab:ablations_ten}).}

\begin{table}[h]

\centering
\small
\caption{\textbf{Search ablations, ten seeds per task.} Fully completed tasks and mean accepted effects. The first block searches joint-action proposals pulled back through the Jacobian, and the second searches the latent command directly.}
\label{tab:ablations_ten}
\begin{tabular}{@{}lccc@{}}
\toprule
Search & Shutter (5) & Hurdle (10) & Door-cube (6) \\
\midrule
\textbf{CEM-LJS} & \textbf{7/10}, 3.9 & \textbf{3/10}, 6.2 & \textbf{10/10}, 6.0 \\
MPPI LJS & 4/10, 2.0 & 2/10, 6.3 & 9/10, 5.4 \\
LATTICE-style LJS & 3/10, 1.5 & 1/10, 3.6 & 7/10, 4.5 \\
Zero-mean refinement LJS & 2/10, 1.0 & 1/10, 4.4 & 2/10, 2.1 \\
Targeted-mean refinement LJS & 1/10, 0.5 & 0/10, 3.6 & 4/10, 4.2 \\
Gaussian-only LJS & 0/10, 0.1 & 1/10, 5.5 & 2/10, 3.0 \\
\midrule
CEM direct-$z$ & 0/10, 0.1 & 1/10, 4.6 & 6/10, 4.9 \\
MPPI direct-$z$ & 3/10, 1.5 & 0/10, 2.3 & 6/10, 4.8 \\
Gaussian direct-$z$ & 0/10, 0.0 & 0/10, 2.7 & 4/10, 4.4 \\
Targeted refinement direct-$z$ & 0/10, 0.0 & 0/10, 1.6 & 3/10, 3.6 \\
Random 29-D latent subspace & 0/10, 0.0 & 0/10, 0.8 & 0/10, 1.9 \\
\bottomrule
\end{tabular}
\end{table}

{\textbf{The coupling helps across optimizers.} CEM-LJS completes each task most often. With the same optimizer (CEM, MPPI, Gaussian, or targeted refinement) on the same task, proposals pulled back through the Jacobian complete more effects than proposals in the latent space in 11 of the 12 pairs. The one exception is the Gaussian search on door-cube. The difference is largest with CEM and smaller with MPPI. A random 29-dimensional latent subspace, which matches the dimension of the LJS proposals, completes no task.}

{\textbf{The optimizer still matters within LJS.} Among the LJS variants, CEM completes the most tasks, and MPPI comes close on hurdle and door-cube. Variants that rely on Gaussian proposals or on targeted refinement complete far fewer tasks, especially on the shutter, where the hands have to be placed precisely beneath the panel.}

\subsection{Sensitivity to the proposal settings}
\label{app:settings_sweeps}

{To check that the comparison does not hinge on the proposal settings of each method, we rerun the main methods on the shutter, where they differ most, with their proposal scale lowered and raised, and CEM-LJS also with its damping $\lambda$ lowered and raised. Everything else follows Appendix~\ref{app:ten_seed} (Table~\ref{tab:settings_sweeps}). Neither baseline gets through the shutter at any of its settings, while CEM-LJS completes it at every setting, most often at its default setting and with the damping quartered. Almost all of its failures at the other settings stop at the first effect, where both palms must be placed beneath the panel. The damping explains much of this. Along the CEM-LJS shutter runs, the policy Jacobian has a median singular value of 0.048, and the damped inverse realizes at least half of a proposed joint change along a median of 24 of the 29 directions with the default $\lambda=0.02$ and 26 with $\lambda\times0.25$, but only 9 with $\lambda\times4$. With four times the damping, most of a proposed joint change no longer reaches the command, and the search completes 2 of 10 shutters, the same as with half the proposal scale.}

\begin{table}[h]

\centering
\small
\caption{\textbf{Shutter with changed proposal settings}, ten seeds per row. Fully completed tasks and mean accepted effects. The first row of each block is the default setting.}
\label{tab:settings_sweeps}
\begin{tabular}{@{}llc@{}}
\toprule
Search & Setting & Shutter (5) \\
\midrule
\textbf{CEM-LJS} & default & \textbf{7/10}, 3.9 \\
 & scale $\times0.5$ & 2/10, 1.2 \\
 & scale $\times2$ & 5/10, 2.9 \\
 & damping $\lambda\times0.25$ & 7/10, 3.5 \\
 & damping $\lambda\times4$ & 2/10, 1.0 \\
\midrule
CEM direct-$z$ & default & 0/10, 0.1 \\
 & scale $\times0.3$ & 0/10, 0.0 \\
 & scale $\times3$ & 0/10, 0.1 \\
\midrule
CEM residual-BFM & default & 0/10, 1.5 \\
 & scale $\times0.5$ & 0/10, 1.2 \\
 & scale $\times2$ & 0/10, 1.0 \\
\bottomrule
\end{tabular}
\end{table}

\subsection{Latent proposal scale, PPO, and SONIC on a single shutter seed}
\label{app:ablations_seed0}

{On a single shutter seed, with the first four shutter effects and a total budget of 1,310,720 candidate control steps, we also test the scale of direct latent proposals, policy learning with PPO, and SONIC as the pretrained controller (Table~\ref{tab:ablations_seed0}).}

\begin{table}[h]
    \centering
    \caption{\textbf{PPO and SONIC on a single shutter seed, first four effects.} {CEM-LJS with BFM-Zero on the same seed is given for reference.} Search cost counts simulated candidate control steps, and successful searches stop early. PPO counts training control steps and is not directly comparable.}
    \label{tab:ablations_seed0}
    \small
    \begin{tabular}{lcc}
        \toprule
        Variant & Effects & Control steps \\
        \midrule
        \textbf{CEM-LJS} (BFM-Zero) & \textbf{4/4} & \textbf{696,320} \\
        \midrule
        PPO direct-$z$ & 0/4 & 1,310,720 training \\
        PPO through LJS & 0/4 & 1,310,720 training \\
        \midrule
        Mixed-Gaussian LJS with SONIC & 1/4 & 696,320 \\
        \bottomrule
    \end{tabular}
\end{table}

{\textbf{Latent proposal scale.}} Direct latent Gaussian scales of $0.15$, $0.5$, and $1.5$ all fail at hand placement under matched budgets, indicating that perturbation magnitude alone does not resolve the precision problem.

{\textbf{Using the score as an RL reward.}} {PPO trained with the negative physical-effect score completes no effect within this budget, either through direct latent commands or through LJS-coupled proposals.} The Jacobian mapping alone therefore does not make this objective easy to optimize through policy learning. {These training interactions differ from candidate-rollout evaluations, so the two budgets are not directly comparable.}

{\textbf{SONIC as the pretrained controller.}} {Replacing BFM-Zero with SONIC \citep{SONIC} does not get beyond the first effect. We tried quantized FSQ tokens and continuous decoder inputs, mixed-Gaussian and CEM search, several noise scales and damping values, and longer horizons. The best run raises the shutter to $0.415\,\mathrm{m}$ of the required $0.75\,\mathrm{m}$, collapses into a low, unstable posture, and never crosses. SONIC tracks supplied motions very well, but in our experiments its inputs could not be steered toward this interaction from the physical-effect scores.}

\section{Motion quality}
\label{app:motion_quality}

{\textbf{Measurements.} We replay every recorded control step of the ten-seed runs in MuJoCo with BFM-Zero's PD law, the joint position and velocity limits of the Unitree G1, and the motor torque limits of the scene. Human-likeness is the logit of BFM-Zero's own motion discriminator, with the latent inferred from each run's states. Neither method optimizes it. We compare CEM-LJS with CEM residual-BFM over the 65 effects that both complete on the same task and seed, using medians and a two-sided Wilcoxon signed-rank test (Table~\ref{tab:motion_quality}). CEM-LJS is better on 12 of the 13 measures. No run of any method exceeds a G1 joint velocity limit. The commanded torque does peak above the motor limit for every method (median peak of 1.3 to 2.1 times the limit per task). Because this also happens for CEM direct-$z$, which only changes the latent command of BFM-Zero, the peaks come from BFM-Zero's own PD tracking during contact rather than from LJS. Whenever the commanded torque exceeds a motor's limit, the motor simply applies its maximum torque instead, as a real motor would. Over the 75 effects that both complete in the runs with retries (Appendix~\ref{app:robustness}), CEM-LJS is better on all 13 measures and uses about half the mechanical power.}

\begin{table}[h]

\centering
\small
\caption{\textbf{Motion quality of CEM-LJS and CEM residual-BFM} over the 65 effects that both complete on the same task and seed (medians). The last two columns give the number of effects on which CEM-LJS is better and the Wilcoxon signed-rank $p$-value.}
\label{tab:motion_quality}
\setlength{\tabcolsep}{4pt}
\begin{tabular}{@{}lcccc@{}}
\toprule
Measure (per effect) & CEM-LJS & \stackhead{CEM\\residual-BFM} & \stackhead{CEM-LJS\\better} & $p$ \\
\midrule
Mechanical power (W/kg) & \textbf{1.59} & 2.61 & 56/65 & $6\times10^{-7}$ \\
Joint jerk, RMS (rad/s$^3$) & \textbf{782} & 1660 & 54/65 & $1\times10^{-7}$ \\
Peak commanded torque / motor limit & \textbf{1.28} & 2.53 & 54/65 & $4\times10^{-8}$ \\
Time at the motor torque limit (fraction) & \textbf{0.06} & 0.20 & 43/65 & $9\times10^{-3}$ \\
Joint-target rate, RMS (rad/s) & \textbf{2.23} & 4.03 & 55/65 & $1\times10^{-7}$ \\
Peak base angular speed (rad/s) & \textbf{2.93} & 4.05 & 58/65 & $3\times10^{-10}$ \\
Peak joint speed / G1 limit & \textbf{0.33} & 0.50 & 54/65 & $5\times10^{-8}$ \\
Excursion beyond joint range (rad) & \textbf{0.005} & 0.025 & 55/65 & $1\times10^{-9}$ \\
Foot slip under load (m/s) & \textbf{0.020} & 0.112 & 55/65 & $2\times10^{-8}$ \\
Fraction of time with self-contact & \textbf{0.24} & 0.42 & 44/65 & $2\times10^{-4}$ \\
Deepest interpenetration (m) & \textbf{0.008} & 0.011 & 41/65 & $2\times10^{-2}$ \\
Peak contact force (body weights) & 3.03 & 3.47 & 35/65 & 0.3 \\
Human-likeness (higher is better) & \textbf{$-0.89$} & $-2.40$ & 57/65 & $5\times10^{-10}$ \\
\bottomrule
\end{tabular}
\end{table}

{\textbf{Shaping the motion within the search.} Since a task is just a set of physical-effect scores, we can add a cost for motion quality to the scores without changing anything else. We add the excess adversarial cost of BFM-Zero's motion discriminator as a cost-only term to every hurdle effect. It never enters the acceptance conditions. We run the search with retries (Appendix~\ref{app:robustness}) on ten seeds, with and without this term, and compare the 79 effects completed in both (Table~\ref{tab:motion_prior}). The crossing becomes calmer and more human-like on most effects. The search becomes longer, and two of the ten seeds reach the time limit.}

\begin{table}[h]

\centering
\small
\caption{\textbf{Hurdle with BFM-Zero's discriminator added to the effect scores}, over the 79 effects completed with and without it on the same seed. Ratios are medians of with / without.}
\label{tab:motion_prior}
\setlength{\tabcolsep}{4pt}
\begin{tabular}{@{}lccc@{}}
\toprule
Measure & With / without & Lower with discriminator & $p$ \\
\midrule
Mechanical power & \textbf{0.71} & 54/79 & $1\times10^{-4}$ \\
Joint jerk, RMS & \textbf{0.80} & 49/79 & $7\times10^{-4}$ \\
Joint-target rate, RMS & \textbf{0.84} & 55/79 & $3\times10^{-4}$ \\
Peak base angular speed & \textbf{0.79} & 54/79 & $4\times10^{-4}$ \\
Peak joint speed / G1 limit & \textbf{0.89} & 49/79 & $1\times10^{-2}$ \\
RMS torque / motor limit & \textbf{0.92} & 51/79 & $3\times10^{-2}$ \\
Deepest interpenetration & \textbf{0.80} & 48/79 & $2\times10^{-2}$ \\
\midrule
Human-likeness higher with discriminator & \multicolumn{3}{c}{67/79 effects ($p=2\times10^{-11}$)} \\
Full hurdle completed & \multicolumn{3}{c}{8/10 with, 10/10 without} \\
Median search time & \multicolumn{3}{c}{688\,s with, 461\,s without} \\
\bottomrule
\end{tabular}
\end{table}

\section{Robustness to new starts and varied physics}
\label{app:robustness}

{\textbf{Moved and turned starts.} We start the shutter task from ten configurations in which the robot is both moved (0.05--0.19\,m, drawn uniformly within 0.20\,m) and turned ($-11.2^\circ$ to $+11.8^\circ$, drawn uniformly within $\pm15^\circ$). The joints and the shutter are unchanged, and so are the effect definitions. The final standing effect asks the robot to face through the doorway, which for the default start is the same as its starting heading. Each start uses one new seed and the settings of Appendix~\ref{app:ten_seed} (Table~\ref{tab:moved_starts}). CEM-LJS gets through from 8 of the 10 starts, more than any baseline (Fisher's exact test, $p\leq0.023$ against each baseline).}

\begin{table}[h]

\centering
\small
\caption{\textbf{Shutter from ten moved and turned starts.} Effects completed from each start, the number of fully completed tasks, and the wall-clock search time of successful runs, mean (fastest).}
\label{tab:moved_starts}
\begin{tabular}{@{}lccc@{}}
\toprule
Method & Effects per start & Full shutter & Search time \\
\midrule
\textbf{CEM-LJS} & 0, 5, 5, 5, 5, 1, 5, 5, 5, 5 & \textbf{8/10} & 84\,s (57\,s) \\
CEM direct-$z$ & 0, 0, 5, 0, 0, 0, 0, 0, 2, 0 & 1/10 & 158\,s (158\,s) \\
CEM residual-BFM & 1, 1, 1, 1, 4, 1, 1, 1, 1, 1 & 0/10 & -- \\
Mixed-Gaussian LJS & 0, 0, 0, 0, 0, 5, 5, 0, 0, 0 & 2/10 & 95\,s (93\,s) \\
\bottomrule
\end{tabular}
\end{table}

{\textbf{Retries and randomized physics.} A CEM search can settle in a poor local optimum, for example when a hurdle effect is accepted in a pose that makes the next effect hard to reach. For practical use, we add two changes that apply to every method. First, a failed effect is searched again from the same state with a new seed, and if it fails again, the previous effect is searched again so that the failing effect starts from a different state, up to six times per run. Second, every candidate is scored under four physics settings drawn from ranges used in humanoid sim-to-real work \citep{he2025asap, yu2026ssr}: friction between 0.2 and 1.1, link masses within $\pm10\%$, PD gains between $-7.5\%$ and $+5\%$ (P) and within $\pm10\%$ (D), and a control delay of 20 to 40\,ms. Each run is limited to 15 minutes. We use ten new seeds, 20 to 29, and each method keeps its own proposal settings and the budgets of Appendix~\ref{app:ten_seed} (Table~\ref{tab:retries}). CEM-LJS completes all 30 runs, while every baseline fails on some of them (Fisher's exact test, $p\leq0.005$ against each baseline). The retries largely remove the factor of luck from a single search, and the remaining trade-off is between robustness and wall-clock time.}

\FloatBarrier
\section{Sanity check with minimal definitions}
\label{app:sanity_check}

{\textbf{Task.} The humanoid stands with a 0.20\,m, 3\,kg box resting on a thin post between its palms. The controller does not squeeze an object on its own, and without the post the box falls within about 0.1\,s, so the search has to make the robot grip the box, lift it off the post, and carry it while walking. The task has ten effects, one for each point of a path. Effect $k$ has a single requirement, $\|p_{\mathrm{box}}-q_k\|_2\leq 0.05\,\mathrm{m}$, scored by the squared distance between the box center and the point $q_k$ (Appendix~\ref{app:carry_math}). One constraint holds throughout every effect: the box never touches the floor. Nothing else is measured, so there are no contact, force, or posture terms that could be tuned toward a method.}

\begin{figure}[h]
\centering
\includegraphics[width=\linewidth]{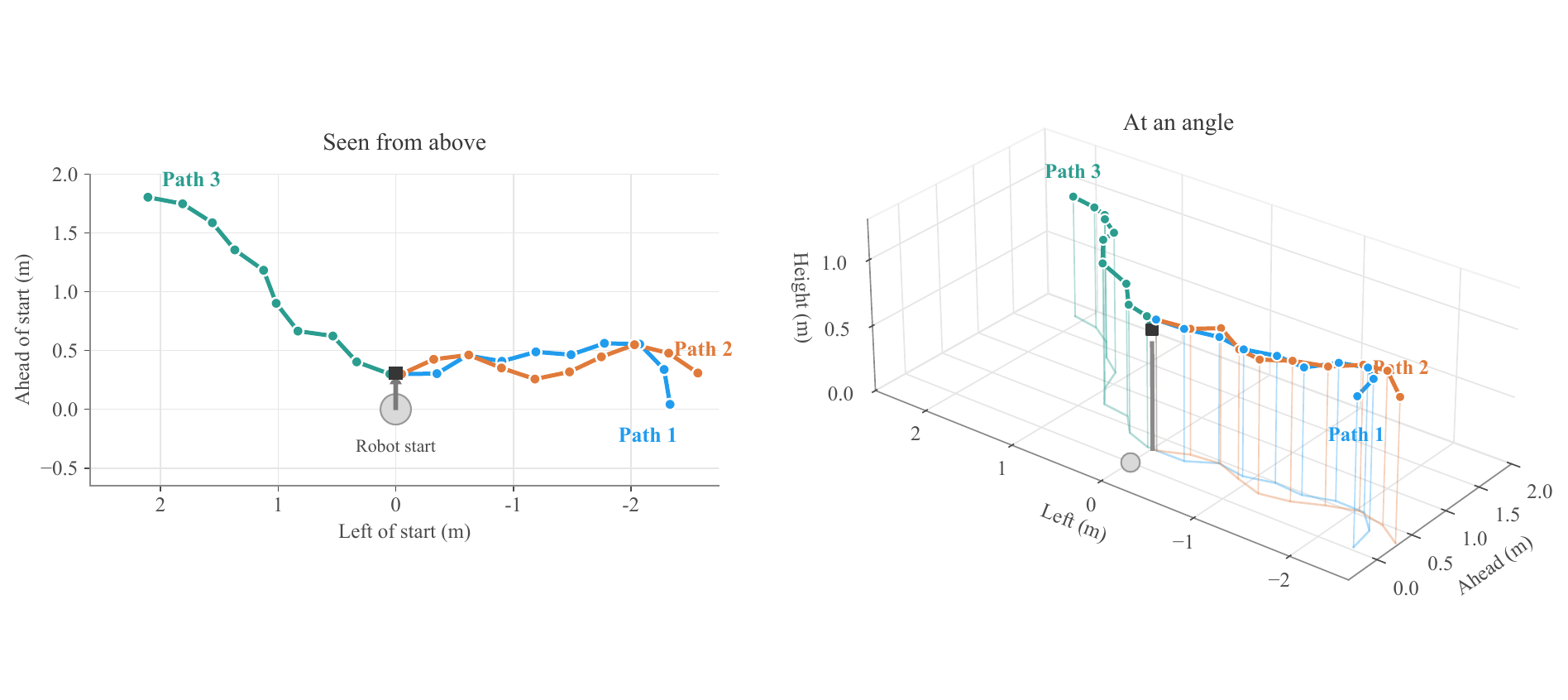}
\caption{\textbf{The three generated paths} for the sanity check, seen from above and at an angle.}
\label{fig:carry_paths_full}
\end{figure}

{\textbf{Paths.} We generate three paths with a fixed rule. Each has ten points, 0.30\,m apart horizontally, at box-center heights between 0.94 and 1.18\,m. The first point lifts the box off the post, after which the path heads to the left or right and turns by up to $45^\circ$ at each point. Paths that come back near the post, cross themselves, turn less than $90^\circ$ in total, or vary less than 0.15\,m in height are drawn again. None of these paths was used while writing the task (Figure~\ref{fig:carry_paths_full}).}

\begin{figure}[!tbp]

\centering
{\small Path 1\par}\vspace{1pt}
\includegraphics[width=\linewidth]{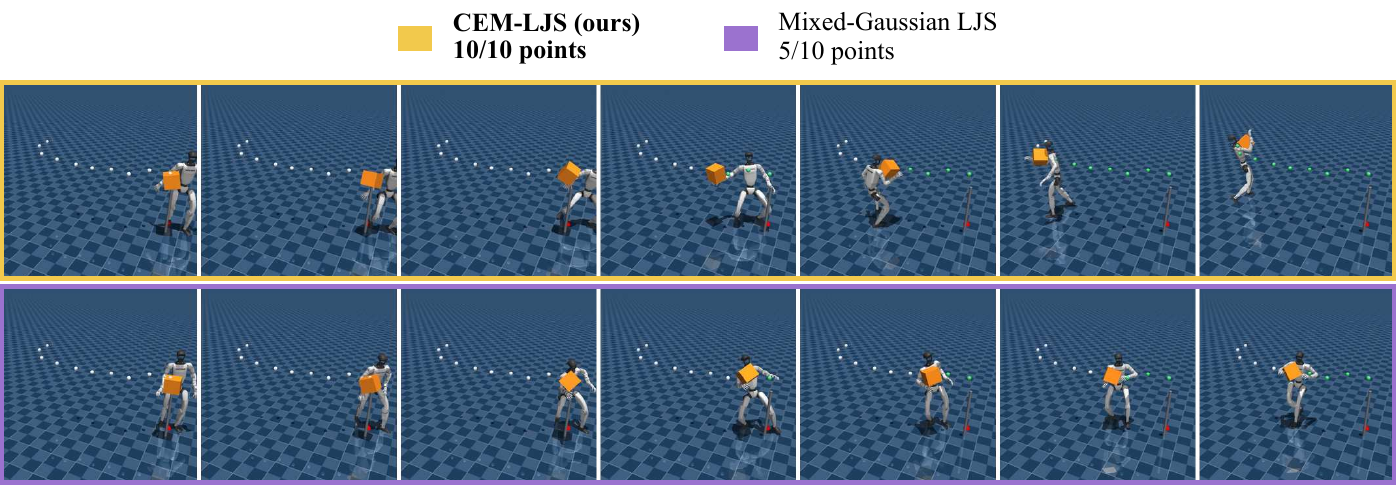}\par\vspace{4pt}
{\small Path 2\par}\vspace{1pt}
\includegraphics[width=\linewidth]{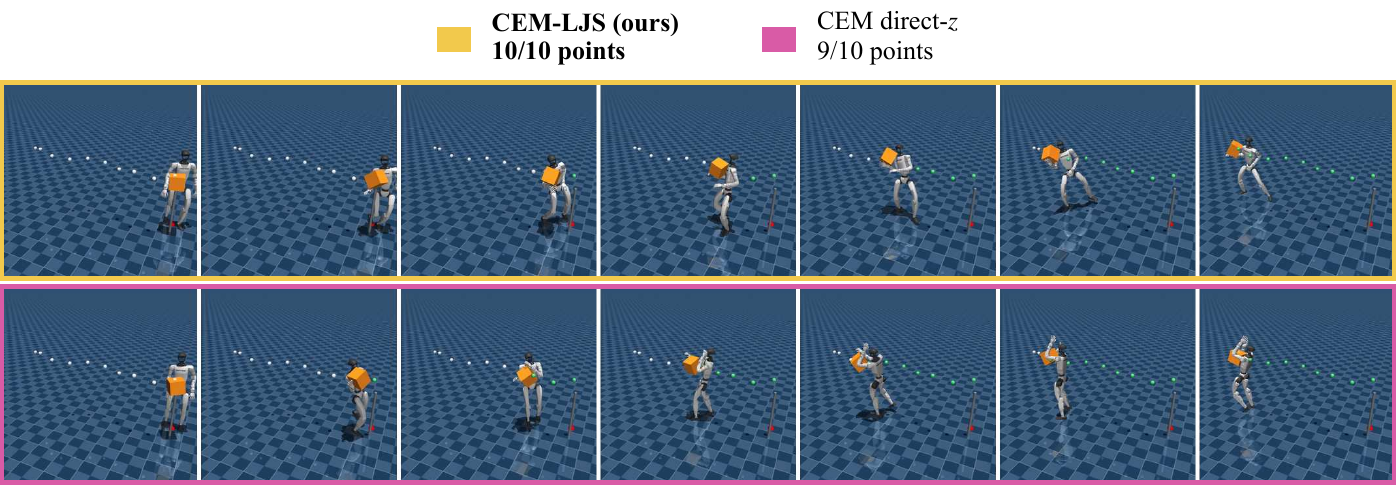}\par\vspace{4pt}
{\small Path 3\par}\vspace{1pt}
\includegraphics[width=\linewidth]{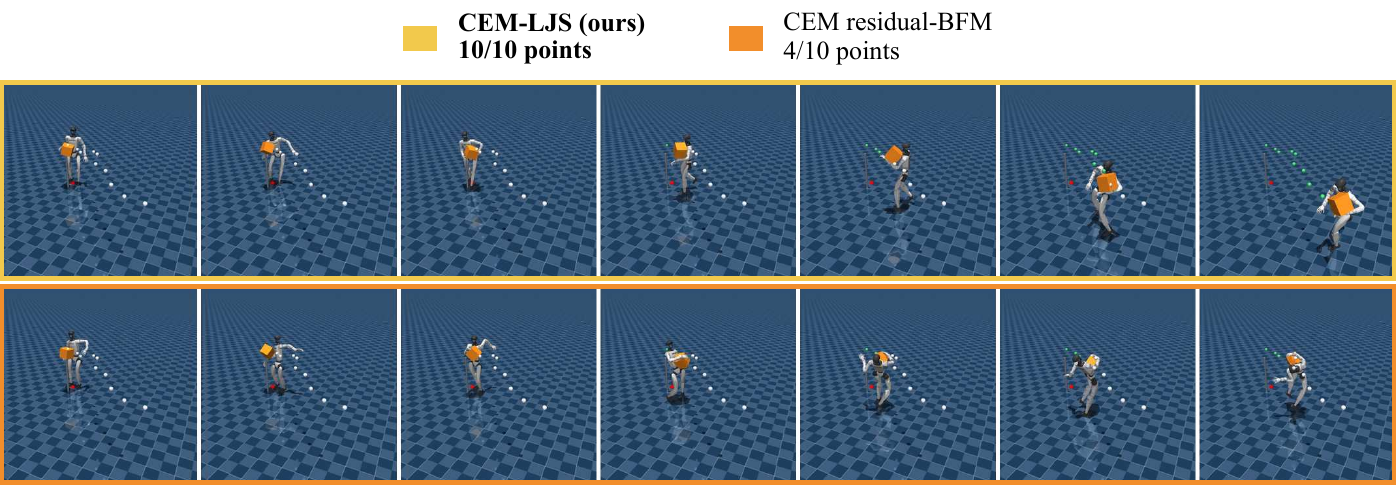}
\caption{\textbf{Sanity check with minimal definitions.} CEM-LJS and one baseline on each path with the same seed, seven evenly spaced frames per run. The small spheres mark the points of the path and turn green once the box reaches them.}
\label{fig:carry_strips}
\end{figure}

{\textbf{Results.} Every method runs five seeds on each path with retries and without randomized physics, a 15-minute limit per run, and the budgets of Appendix~\ref{app:ten_seed}. A run counts as complete when the box reaches all ten points in order without touching the floor, which we check by replaying the recorded states (Table~\ref{tab:sanity_check} and Figure~\ref{fig:carry_strips}). CEM-LJS completes 12 of the 15 runs, against at most 3 for any baseline (Fisher's exact test, $p\leq0.003$ against each baseline).}

\begin{table}[h]

\centering
\small
\caption{\textbf{Sanity check with minimal definitions.} Completed runs per path (five seeds each) and the mean number of points reached.}
\label{tab:sanity_check}
\begin{tabular}{@{}lccccc@{}}
\toprule
Method & Path 1 & Path 2 & Path 3 & All & Points reached (of 10) \\
\midrule
\textbf{CEM-LJS} & \textbf{4/5} & \textbf{4/5} & \textbf{4/5} & \textbf{12/15} & \textbf{9.1} \\
CEM direct-$z$ & 1/5 & 1/5 & 0/5 & 2/15 & 2.5 \\
CEM residual-BFM & 2/5 & 0/5 & 1/5 & 3/15 & 4.8 \\
Mixed-Gaussian LJS & 1/5 & 0/5 & 0/5 & 1/15 & 1.5 \\
\bottomrule
\end{tabular}
\end{table}

\section{Assembly with the frozen Play2Perfect policy}
\label{app:play2perfect}

{\textbf{Controller and tasks.} Play2Perfect \citep{lum2026play2perfect} pretrains a goal-conditioned policy for a Sharpa hand on a Franka arm through play, and then fine-tunes it separately for each assembly task. We use only the pretrained play policy, without any fine-tuning, and evaluate it in MuJoCo on the four assembly tasks of Play2Perfect. The effects are: pick up, mate, and release for the first beam part, pick up, approach, mate, and release for the second beam part, pick up, engage, turn, and release for screwing, and pick up, seat, and release for the tight insertion. Screwing uses a different policy mode of the same checkpoint.}

\begin{figure}[p]

\centering
{\small Beam, first part\par}\vspace{1pt}
\includegraphics[width=\linewidth]{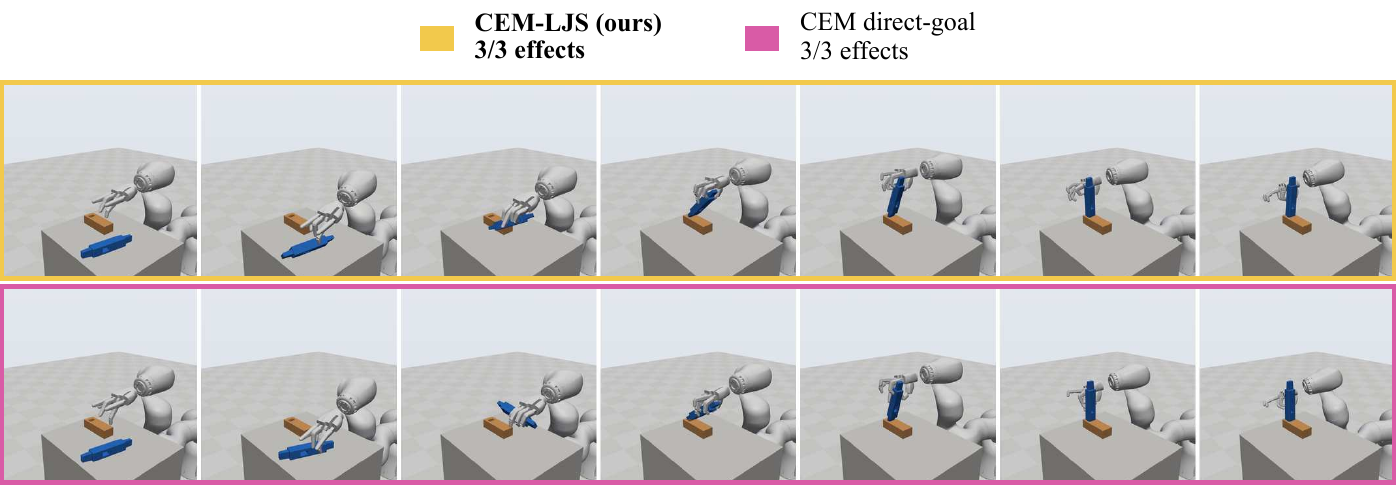}\par\vspace{4pt}
{\small Beam, second part\par}\vspace{1pt}
\includegraphics[width=\linewidth]{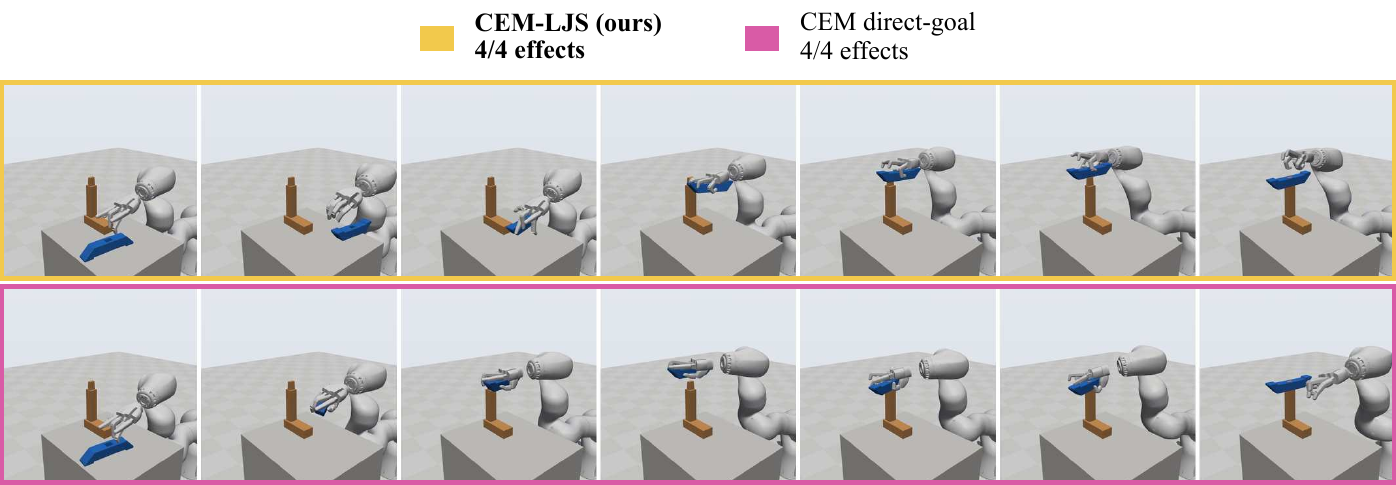}\par\vspace{4pt}
{\small Screwing\par}\vspace{1pt}
\includegraphics[width=\linewidth]{figures/screwing_comparison.pdf}\par\vspace{4pt}
{\small Insertion\par}\vspace{1pt}
\includegraphics[width=\linewidth]{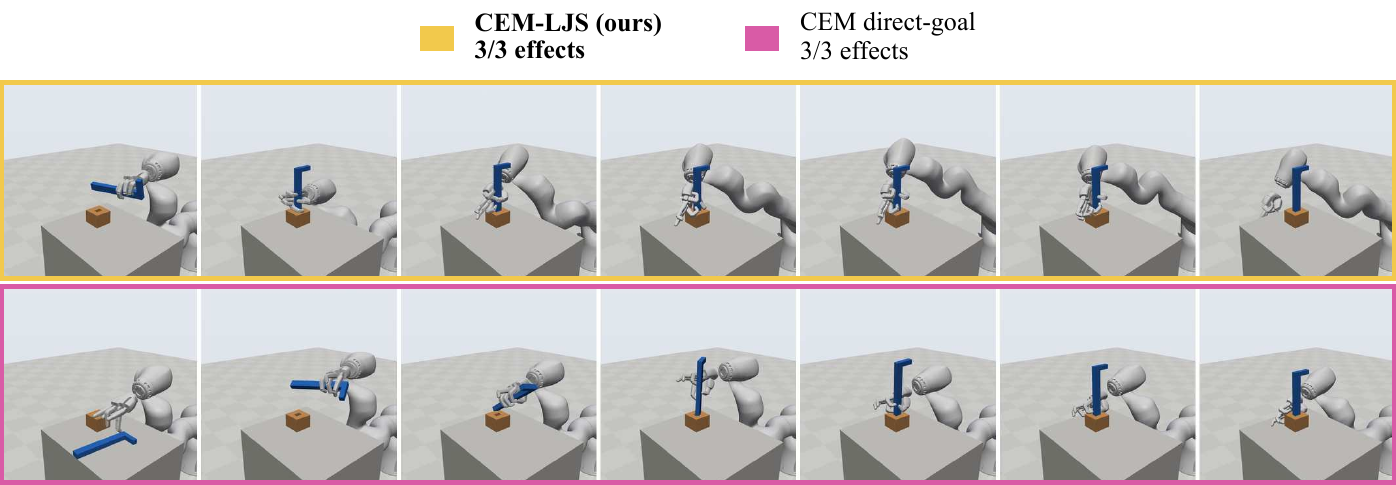}
\caption{\textbf{Assembly with the frozen Play2Perfect policy.} Seven evenly spaced frames from a successful run of each method on each task.}
\label{fig:p2p_comparison}
\end{figure}

{\textbf{Search.} CEM-LJS proposes changes to the 29 joint targets and maps them to the 6-dimensional goal through the damped inverse of the policy Jacobian. CEM direct-goal searches the six translation and rotation coordinates of the goal directly, with initial standard deviations of 0.1\,m and 0.5\,rad. Both use the same physical-effect scores, goal bounds, and elite selection, 1,024 candidates per decision (4,096 for the first beam part), a 20-step prediction, and a 10-step execution. The ten seeds were not used while defining the effects. Over the four tasks, CEM-LJS completes 31 of 40 runs against 21 for CEM direct-goal (Fisher's exact test, $p=0.034$), and the largest gap is on screwing (9 against 3, $p=0.020$). Table~\ref{tab:p2p_times} gives the completed tasks with their search times, Figure~\ref{fig:p2p_search_cost} the search cost to reach each effect, Figure~\ref{fig:p2p_comparison} frames from the runs, and Table~\ref{tab:seed_results_p2p} the individual runs.}

\begin{table}[h]

\centering
\small
\caption{\textbf{Assembly with the frozen Play2Perfect policy, ten seeds.} Fully completed tasks, and the wall-clock search time of successful runs, mean (fastest), on an NVIDIA RTX 4090.}
\label{tab:p2p_times}
\begin{tabular}{@{}lcccc@{}}
\toprule
Task & CEM-LJS & CEM direct-goal & Time, CEM-LJS & Time, CEM direct-goal \\
\midrule
Beam, first part & \textbf{10/10} & \textbf{10/10} & 62\,s (29\,s) & 46\,s (29\,s) \\
Beam, second part & \textbf{7/10} & 4/10 & 72\,s (43\,s) & 65\,s (48\,s) \\
Screwing & \textbf{9/10} & 3/10 & 93\,s (52\,s) & 132\,s (88\,s) \\
Insertion & \textbf{5/10} & 4/10 & 83\,s (69\,s) & 64\,s (22\,s) \\
\midrule
All & \textbf{31/40} & 21/40 & & \\
\bottomrule
\end{tabular}
\end{table}

\begin{table}[h]

\centering
\scriptsize
\setlength{\tabcolsep}{3pt}
\caption{\textbf{Wall-clock search time and outcome of every Play2Perfect evaluation run, ten seeds.} Each entry gives the accepted effects / task total and, in parentheses, the search time in seconds on an NVIDIA RTX 4090. Fully completed tasks are in bold.}
\label{tab:seed_results_p2p}
\resizebox{\linewidth}{!}{%
\begin{tabular}{@{}llcccccccccc@{}}
\toprule
Task & Method & 929560953 & 875095668 & 1769421694 & 1418357545 & 885480420 & 1284918341 & 747747906 & 2023936781 & 1602613270 & 1370732405 \\
\midrule
Beam, first & CEM-LJS & \textbf{3/3\,(106)} & \textbf{3/3\,(36)} & \textbf{3/3\,(47)} & \textbf{3/3\,(46)} & \textbf{3/3\,(35)} & \textbf{3/3\,(29)} & \textbf{3/3\,(66)} & \textbf{3/3\,(93)} & \textbf{3/3\,(62)} & \textbf{3/3\,(99)} \\
 & CEM direct-goal & \textbf{3/3\,(29)} & \textbf{3/3\,(76)} & \textbf{3/3\,(41)} & \textbf{3/3\,(96)} & \textbf{3/3\,(42)} & \textbf{3/3\,(51)} & \textbf{3/3\,(30)} & \textbf{3/3\,(34)} & \textbf{3/3\,(33)} & \textbf{3/3\,(33)} \\
\midrule
Beam, second & CEM-LJS & 1/4\,(107) & 3/4\,(114) & 1/4\,(98) & \textbf{4/4\,(60)} & \textbf{4/4\,(75)} & \textbf{4/4\,(56)} & \textbf{4/4\,(119)} & \textbf{4/4\,(51)} & \textbf{4/4\,(96)} & \textbf{4/4\,(43)} \\
 & CEM direct-goal & 2/4\,(143) & 1/4\,(142) & 1/4\,(92) & 1/4\,(86) & 1/4\,(102) & \textbf{4/4\,(51)} & \textbf{4/4\,(48)} & 1/4\,(130) & \textbf{4/4\,(72)} & \textbf{4/4\,(91)} \\
\midrule
Screwing & CEM-LJS & \textbf{4/4\,(52)} & \textbf{4/4\,(110)} & \textbf{4/4\,(93)} & \textbf{4/4\,(140)} & \textbf{4/4\,(83)} & \textbf{4/4\,(76)} & \textbf{4/4\,(162)} & \textbf{4/4\,(64)} & \textbf{4/4\,(58)} & 2/4\,(156) \\
 & CEM direct-goal & \textbf{4/4\,(88)} & \textbf{4/4\,(155)} & 1/4\,(125) & 1/4\,(147) & 1/4\,(121) & 2/4\,(152) & \textbf{4/4\,(152)} & 0/4\,(114) & 0/4\,(123) & 1/4\,(150) \\
\midrule
Insertion & CEM-LJS & \textbf{3/3\,(106)} & \textbf{3/3\,(76)} & \textbf{3/3\,(78)} & \textbf{3/3\,(88)} & 1/3\,(116) & 1/3\,(132) & 1/3\,(126) & 1/3\,(136) & \textbf{3/3\,(69)} & 1/3\,(110) \\
 & CEM direct-goal & 1/3\,(112) & 1/3\,(125) & \textbf{3/3\,(35)} & 1/3\,(128) & \textbf{3/3\,(95)} & 1/3\,(127) & \textbf{3/3\,(103)} & 1/3\,(124) & \textbf{3/3\,(22)} & 1/3\,(129) \\
\bottomrule
\end{tabular}}
\end{table}

\section{Other controllers and simulators}
\label{app:other_controllers}

\begin{figure}[h]

\centering
\includegraphics[width=\linewidth]{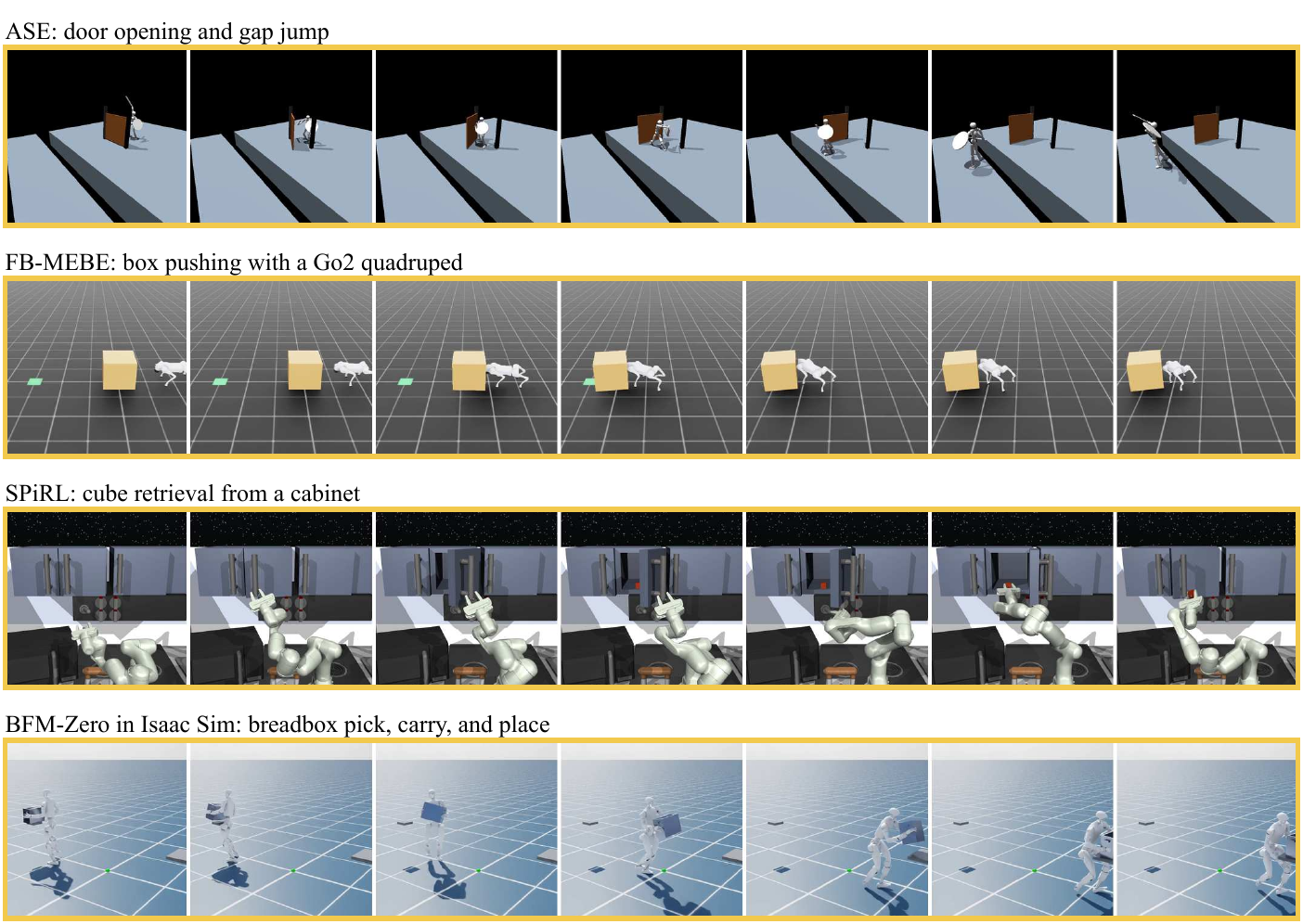}
\caption{\textbf{LJS with further pretrained controllers and a second simulator.} Seven evenly spaced frames from one search per controller.}
\label{fig:other_controllers}
\end{figure}

{LJS needs only a frozen policy that takes a latent or a goal, and a simulator to evaluate candidates. To show how easily it carries over to any steerable latent- or goal-conditioned pretrained controller, we also steer the following controllers, embodiments, and simulators. Figure~\ref{fig:other_controllers} shows frames from these searches, and their videos are available on the project page.}

{\textbf{ASE} \citep{peng2022ase}, in Isaac Gym. With the frozen sword-and-shield controller, LJS opens a passive door, jumps across a gap and lands stably, and in a second task opens a swing gate through contact, walks through it, and stops upright.}

{\textbf{FB-MEBE} \citep{FBMEBE}, in Isaac Sim. With the frozen policy of a Unitree Go2 quadruped, LJS pushes a passive box to a target and stops it there.}

{\textbf{SPiRL} \citep{SPiRL}, in the MuJoCo kitchen. Through the frozen skill decoder, LJS opens a cabinet, grasps a cube and pulls it out, carries it under the door, and pushes the door closed.}

{\textbf{BFM-Zero in a second simulator.} In Isaac Sim with PhysX, LJS picks up a breadbox, carries it, and places it, replanning in a separate simulator process while the robot keeps moving.}

\FloatBarrier
\section{A closed-loop policy from one CEM-LJS rollout}
\label{app:closed_loop}

To see whether the behaviors LJS discovers can be used downstream, we trained a closed-loop policy from a single CEM-LJS rollout on a new task. Writing the effects, searching and training took three to four hours in total. This section describes the task, the search, the policy, and how we evaluated it. Videos of the policy, and a version that runs live in the browser, are on the project page.

\textbf{Task.} The table is 0.75\,m tall, 1.45\,m long and 1.0\,m deep, hollow underneath, with its long side against a wall. The G1 starts standing 0.55\,m in front of the near edge. The eight effects ask for (1) both palms on the table top, (2) the right knee on the top, (3) the rest of the right leg on the top, (4) the chest down onto the table, (5) the whole body on the table, (6) a crouch on both feet on the top, (7) a rise onto both feet, and (8) standing up and settling. Throughout the climb, no body part may reach in under the table top, and nothing but the feet may touch the floor. As for the other tasks, an LLM coding agent helped translate these descriptions into scores and acceptance conditions (Section~\ref{sec:defining_scores}).

\textbf{Search.} CEM-LJS found the climb with the frozen BFM-Zero controller, retrying failed effects and scoring every candidate under four randomized physics settings of friction, link masses, PD gains and control delay (Appendix~\ref{app:robustness}). The robot model applies the G1's motor torque limits. The climb was searched in three parts, each continuing from the saved end of the previous one. The resulting climb lasts 34.4\,s.

\textbf{Policy.} SuperMimic \citep{liao2026supermimic} learns a world model of the robot from simulator data, on 8-step windows, and improves the policy by back-propagating through 32-step rollouts imagined in that world model. We used its official code on mjlab with MuJoCo Warp and added the table. Each simulated world has its own randomized table, and the table's near-edge point is part of the world-model state, so that imagined rollouts know where that world's table is. The reference is the searched climb followed by a 1\,s hold, at 50\,Hz. It is worth noting that the search used BFM-Zero's G1 model, while the policy runs on mjlab's G1 model, which has the same kinematics but different collision geometry. The reward is SuperMimic's tracking reward with action-rate and joint-limit penalties.

The actor sees five frames of on-robot measurements, each with uniform noise: the gravity direction in the pelvis frame ($\pm$0.05), the pelvis IMU angular velocity ($\pm$0.2\,rad/s), joint positions with encoder bias ($\pm$0.01\,rad), joint velocities ($\pm$0.5\,rad/s), its last action, and the table's near-edge point in the pelvis frame with a visibility flag ($\pm$3\,cm). The edge point is given only while the edge is more than 0.10\,m ahead of the pelvis, as the head camera would see it. The actor also sees the reference joint targets at 10 future frames over 0.9\,s and the reference pelvis orientation relative to its own. It is given no base linear velocity, no root position, and no state estimate.

The table is redrawn every episode, with its near edge moved by up to $\pm$4\,cm toward the robot, $\pm$8\,cm sideways, its height changed by up to $\pm$3\,cm, and its friction drawn from 0.5 to 1.1. Each simulated world also draws its own robot: PD gains scaled by 0.85--1.15 ($k_p$) and 0.8--1.2 ($k_d$) per actuator, torso mass scaled by 0.9--1.15, a torso centre-of-mass offset of up to $\pm$2.5\,cm forward and $\pm$5\,cm sideways and vertically, foot friction from 0.3 to 1.2, and joint encoder bias. Training used 4,096 parallel worlds on one NVIDIA RTX PRO 6000 GPU and took 45 minutes, about 80 million environment steps.

\textbf{Evaluation.} Each trial starts with the robot standing in front of the table and runs the whole climb in closed loop. A trial succeeds if the climb runs to the end without termination and the pelvis ends at least 0.6\,m above the table top, inside its footprint, and within $20^\circ$ of upright. An episode terminates if the pelvis drifts more than 0.5\,m from the reference or its orientation drifts too far. For the second simulator, we export the actor and run it in standard MuJoCo on the CPU, with the same robot and table model, a 5\,ms physics step and four physics steps per control step. Its observations are built from the simulator's IMU and joint readings with the same rules as in training, and each trial draws the table's placement, height and friction from the training ranges. Table~\ref{tab:closed_loop} gives the results. All 21 failures on randomized tables in the training simulator were terminations for drifting from the reference.

\begin{table}[h]
\centering
\small
\caption{\textbf{Closed-loop policy trained from one CEM-LJS rollout}, climbing onto the table and standing up.}
\label{tab:closed_loop}
\begin{tabular}{@{}llc@{}}
\toprule
Simulator & Table & Success \\
\midrule
MuJoCo Warp (training) & Nominal & 496/512 (96.9\%) \\
MuJoCo Warp (training) & Randomized placement, height and friction & 491/512 (95.9\%) \\
MuJoCo, CPU (no retraining) & Randomized placement, height and friction & 48/48 \\
\bottomrule
\end{tabular}
\end{table}

\FloatBarrier
\section{Qualitative rollout comparisons}
\label{app:qualitative_rollouts}

\begin{figure}[!ht]
  \centering
  \includegraphics[width=\linewidth]{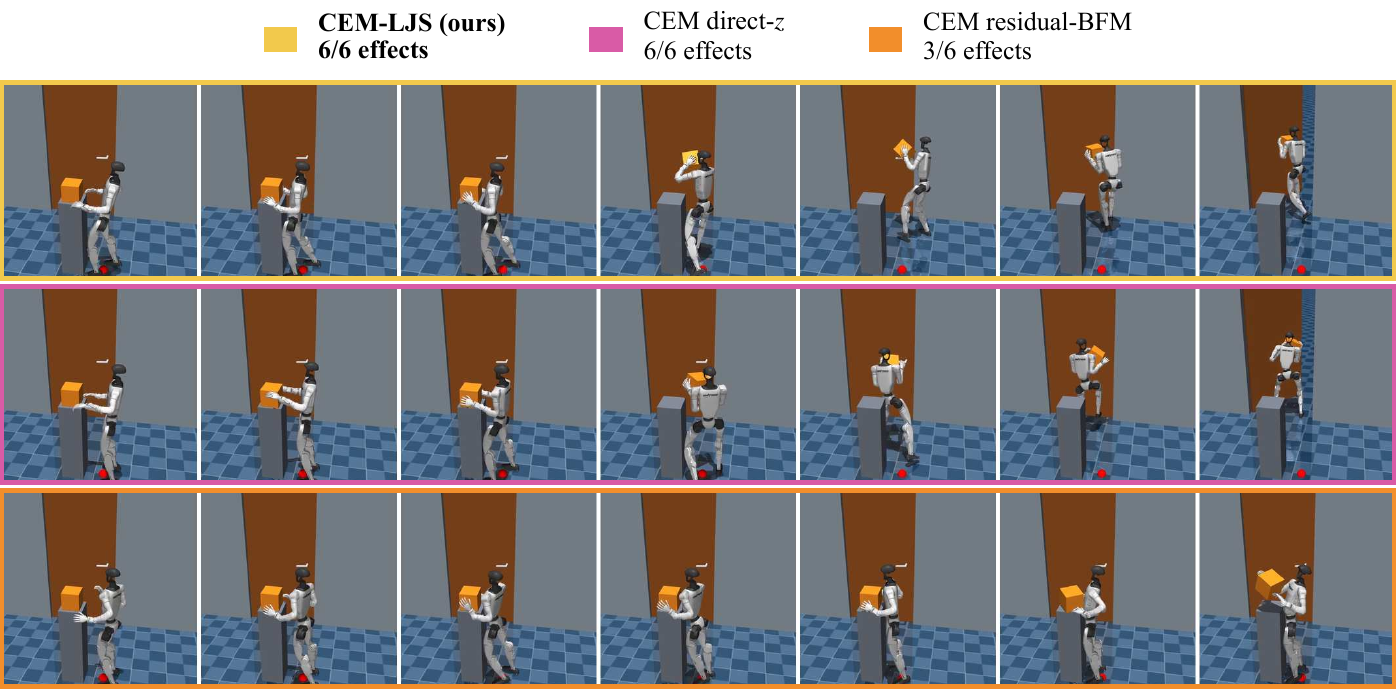}
  \caption{{\textbf{Door opening while carrying a cube.} Seven evenly spaced frames from the best run of each method over the ten seeds. CEM-LJS and CEM direct-$z$ complete all six effects, and CEM residual-BFM completes three.}}
  \label{fig:door_cube_qualitative}
\end{figure}

\begin{figure}[!ht]
  \centering
  \includegraphics[width=\linewidth]{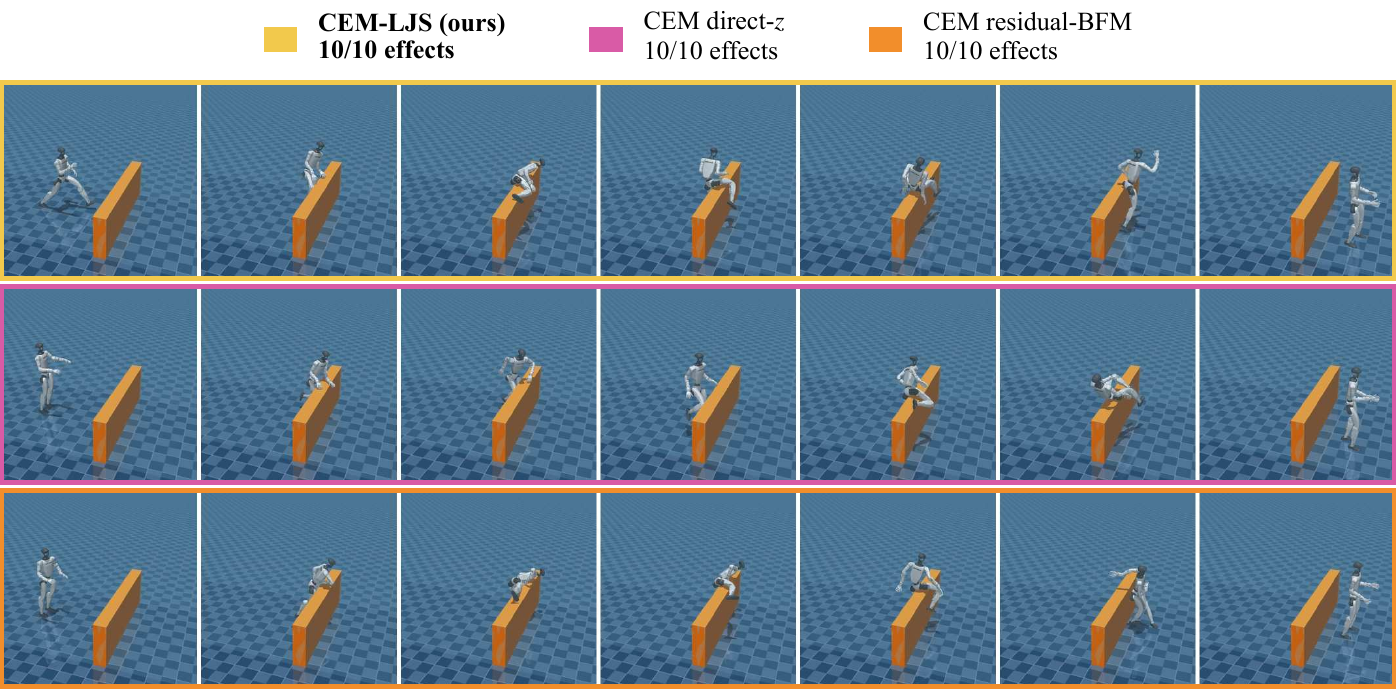}
  \caption{{\textbf{Hurdle traversal.} Seven evenly spaced frames from the best run of each method over the ten seeds. Each of these runs completes all ten effects.}}
  \label{fig:hurdle_qualitative}
\end{figure}

{Figures~\ref{fig:door_cube_qualitative} and~\ref{fig:hurdle_qualitative} compare CEM-LJS with the best run of each baseline over the ten seeds. On door-cube, CEM-LJS and CEM direct-$z$ both open the door with the grasped cube, while CEM residual-BFM lifts the cube but never brings it to the door. On the hurdle, the best run of every method completes the crossing, and Appendix~\ref{app:motion_quality} compares how they move. The videos of all evaluation runs are available on the project page.}

\FloatBarrier
\section{Physical-effect definitions and human--AI score development}
\label{app:effect_definitions}

This appendix reports how the physical-effect definitions were developed and provides the mathematical scores and acceptance conditions for the three evaluated tasks. The score functions are not part of the LJS algorithm. They define the evaluated tasks, and every method is evaluated with the same definitions. The interaction card below follows the compact alternating-contribution format of \citet{feng2026autonomousmathematicsresearch}. The human specified the intended physical outcome, supplied the simulator conventions, corrected geometric and temporal assumptions, and accepted the final definition. The language model translated those instructions into candidate measurement and score code. It did not observe rollouts, choose candidates, control the robot, or modify an effect during a reported search.

\subsection{Human--AI interaction card: first hurdle effect}
\label{app:hurdle_interaction_card}

\begingroup
  \small
  \setlength{\tabcolsep}{5pt}
  \renewcommand{\arraystretch}{1.12}
  \setlength\LTleft{\fill}
  \setlength\LTright{\fill}
  \begin{longtable}{>{\raggedright\arraybackslash}p{0.17\linewidth} >{\raggedright\arraybackslash}p{0.75\linewidth}}
    \caption{\textbf{Human--AI interaction card for the first hurdle effect.} The card records the substantive revisions rather than reproducing the full prompts and code responses. The final scores and acceptance conditions are given in Appendix~\ref{app:hurdle_math}.}
    \label{tab:hurdle_hai_card}\\
    \toprule
    \textbf{Contributor} & \textbf{Contribution} \\
    \midrule
    \endfirsthead
    \multicolumn{2}{l}{\footnotesize\textit{Table~\ref{tab:hurdle_hai_card} continued.}}\\
    \toprule
    \textbf{Contributor} & \textbf{Contribution} \\
    \midrule
    \endhead
    \bottomrule
    \endfoot
    \textbf{Human}
      & Established the coding protocol: define one physical effect at a time for LJS with frozen BFM-Zero; implement measurements, progress scores, and acceptance conditions using the supplied MuJoCo interface; use simple scaled residuals; keep score and acceptance separate; and do not introduce joint-pose commands, inverse kinematics, motion tracking, fine-tuning, teleportation, or external forces. Any revised score must be rerun from the preceding accepted effect. \\
    \textbf{Gemini 3.8 Flash}
      & Restated the protocol and requested the task geometry, entity identifiers, and first-stage instruction before generating code. \\
    \midrule
    \textbf{Human}
      & Specified the fixed hurdle, crossing and lateral axes, right-leg lead, palm-normal conventions, collision-surface geometry, and contact-force convention. Requested the first effect: place both palms on the hurdle top, facing downward, with separate bilateral support, while keeping the robot within the usable width. Either hand could settle anywhere suitable; no exact hand target or extra foot-lifting stage was allowed. \\
    \textbf{Gemini 3.8 Flash}
      & Proposed a complete first-stage module using collision geometry for palm placement, palm-normal alignment, per-hand contact loads, and pelvis centering, together with initial model-chosen tolerances. \\
    \midrule
    \textbf{Human}
      & Corrected three implementation details: use each transformed hand geometry's bounding-box center for horizontal placement and its minimum vertex for height; measure support only at the final frame rather than claiming a sustained five-frame hold; and evaluate all surface bounds in the hurdle's live local frame instead of hardcoding its world center. \\
    \textbf{Gemini 3.8 Flash}
      & Revised the measurements to use those distinct geometric statistics, instantaneous final-frame forces, and the hurdle transform recorded in each frame. \\
    \midrule
    \textbf{Human}
      & Refined the intended outcome after inspecting the definition: keep both hands near the middle to leave room for the next leg movement, but avoid exact positions; require only reasonably downward palms and light bilateral load; reject side-friction support; keep the robot roughly centered; and stop penalizing small placement errors once the palms are suitably on top. Valid near-edge top contact had to count, so top support was to be identified from contact position and normal direction rather than an arbitrary edge strip. \\
    \textbf{Gemini 3.8 Flash}
      & Added placement and orientation deadbands, a centered admissible band, light per-hand support targets, and top-face contact filtering using both local contact position and normal alignment. The human reviewed this proposal before the definition was incorporated into the finalized hurdle module. \\
  \end{longtable}
\endgroup

\noindent\footnotesize\textbf{Raw prompts and outputs:} \url{https://share.gemini.google/GgqoaJR2Yw4F}
\normalsize

The successive corrections are part of the task-specification process: they remove assumptions that were inconsistent with the simulator semantics or with the intended physical outcome. The final reported searches use fixed, human-reviewed definitions.

\subsection{Mathematical physical-effect definitions}
\label{app:effect_code}
\label{app:task_math}

We give the scores and acceptance conditions for the ten hurdle effects, {five} shutter effects, and six door-cube effects{, and the ten effects of the sanity check}. Distances, forces, and times are in metres, newtons, and seconds. Angles are in radians unless degrees are shown. Measurements are taken at the final frame of the effect history $\xi$, except where a temporal minimum or maximum is specified.

\paragraph{Residual and acceptance conventions.}
Define elementwise residuals
\[
 L_a(v;b)=\frac{[b-v]_+}{a},\qquad
 U_a(v;b)=\frac{[v-b]_+}{a},\qquad
 N_{a,d}(v;b)=\frac{[|v-b|-d]_+}{a}.
\]
Vector scales and tolerances are applied componentwise. For each effect we list residual blocks $r_{k,j}$; its score is
\[
 c_k(\xi)=\sum_j\|r_{k,j}(\xi)\|_2^2.
\]
By default, acceptance requires every residual block to vanish. A superscript $\mathrm{g}$ marks a guidance-only block excluded from acceptance. Explicit acceptance overrides below replace the zero-residual test for the named block. Requirements marked $\mathrm{c}$ additionally reject a candidate when their acceptance test fails on the full predicted effect segment. The scores are evaluated over successive trajectory prefixes and averaged over the latter half of the prediction horizon, as in the main method.

\subsubsection{Shutter traversal}
\label{app:shutter_math}
Let $\mathcal{P}_G(s)$ be the transformed collision vertices of body group $G$, with $G=U$ for the head and torso and $G=R$ for the entire robot. The shutter's lower edge and far face are $y_b(s)$ and $x_f(s)$. Define
\[
 d_G(s)=\min_{p\in\mathcal{P}_G(s)}p_x-x_f(s),\qquad
 a_U(s)=\left[\max_{\substack{p\in\mathcal{P}_U(s)\\p_x\leq x_f(s)+.02}}p_z+.02-y_b(s)\right]_+,
\]
where $a_U=0$ if the restricted vertex set is empty. For each hand, its placement point uses the mean vertex coordinates in $x,y$ and the maximum vertex height. Let $e_h\in\mathbb{R}^{6}$ stack its displacement from the two targets $(x_{\mathrm{panel}},y_{\mathrm{panel}}\pm.20,y_b-.005)$. Let $n_h\in\mathbb{R}^2$ contain the upward components of the palm normals. Finally, $F_a$ is the signed net upward force on the panel from hand, wrist, and elbow contacts.

\begin{enumerate}
\item \textbf{Palms beneath the shutter.}
\[
 r_1=\operatorname{col}\bigl(N_{.10,.04}(e_h;0),\ L_1(n_h;\cos30^\circ)\bigr).
\]
Acceptance instead permits $\|e_h\|_\infty\leq.10$ and both upward components at least $\cos45^\circ$.
\item \textbf{Lift the shutter.}
\[
 r_2=\operatorname{col}\bigl(L_{.10}(y_b;.75),\ L_{10}(F_a;1)\bigr).
\]
Acceptance requires a lower-edge height of at least $.75$ and at least $1$ N of upward arm contact.
\item \textbf{Head and torso through.}
\[
 r_3=\operatorname{col}\bigl(U_{.10}(a_U;0),\ L_{.10}(d_U;.02)\bigr).
\]
\item \textbf{Whole body clear.}
\[
 r_4=L_{.10}(d_R;.02).
\]
\item {\textbf{Stand on the far side.}
Let $p^\star=(1.8,0,z_{\mathrm{nom}})$, where the horizontal point $(1.8,0)$ lies $.565$ m beyond the far face on the centre line of the opening and $z_{\mathrm{nom}}$ is the root height of the nominal standing pose. Let $e_{\mathrm{rot}}$ be the quaternion rotation difference from the identity orientation, which faces $+x$ through the opening. With nominal joint positions $q_{\mathrm{nom}}$ and root speed $v=\|\dot p_{\mathrm{root}}\|_2$, define
\[
 e_{\mathrm{stand}}=\operatorname{col}\bigl((p_{\mathrm{root}}-p^\star)/.10,\ e_{\mathrm{rot}}/.5,\ q_{\mathrm{joints}}-q_{\mathrm{nom}}\bigr),
 \qquad
 r_5=\operatorname{col}\bigl(e_{\mathrm{stand}}^{\mathrm{g}},\ U_1(v;.10)\bigr).
\]
Acceptance requires at least $.5$ s of history and the following over the final $.5$ s: $\min_s(R_{\mathrm{torso}}(s))_{zz}\geq\cos10^\circ$, $\min_s d_R(s)\geq.02$, $\min_s z_{\mathrm{root}}(s)\geq.9\,z_{\mathrm{nom}}$, $\max_s v(s)\leq.20$, and a floor normal load of at least $1$ N on each foot in at least $80\%$ of the frames.}
\end{enumerate}

\subsubsection{Door opening while carrying a cube}
\label{app:door_cube_math}
Let $p_c,R_c,a$ denote the cube centre, rotation, and half-extents. The palm contact point is the collision-mesh mean shifted to the inward-most palm surface along the hand's local inward axis. For sides $\epsilon\in\{+1,-1\}$, define stacked palm errors relative to cube-local targets
\[
 e_{\mathrm{reach},\epsilon}=p_{h,\epsilon}-p_c-R_c(a_x+.03,\epsilon(a_y+.02),0)^\top,
 \qquad
 e_{\mathrm{grasp},\epsilon}=p_{h,\epsilon}-p_c-R_c(.045,\epsilon a_y,0)^\top.
\]
Let $v_f\in\mathbb{R}^2$ be the dot products between normalized elbow-to-palm directions and the cube's forward axis. Let $b=(R_c^\top(p_c-p_{\mathrm{torso}}))_{xy}$, $u=(R_{\mathrm{torso}})_{zz}$, and $v$ be the maximum root linear speed over the last $.10$ s. The maximum cube translation and rotation relative to its pose at the beginning of the effect are
\[
 m_p=\max_{s\in\xi}\|p_c(s)-p_c(s_0)\|_2,\qquad
 m_R=\max_{s\in\xi}\arccos\!\left(\frac{\operatorname{tr}(R_c(s_0)^\top R_c(s))-1}{2}\right),
\]
with the arccos argument clipped to $[-1,1]$. Define
\[
 r_m=U_{(.06,15^\circ)}\bigl((m_p,m_R);(.02,5^\circ)\bigr),\qquad
 h_c=p_{c,z}-\sum_j |R_{c,zj}|a_j,\qquad h_s=h_c-z_{\mathrm{stand}}.
\]
Here $z_{\mathrm{stand}}$ is the stand's top height. $F_s$ is the sum of positive cube--stand normal loads. A bar below a measurement denotes its minimum over the final $.08$ s; a bar above denotes its maximum over that window.

\paragraph{Contact and separation measurements.}
$F_h\in\mathbb{R}^2$ contains the minimum inward palm loads over $.08$ s. For hand $\epsilon$, a contact contributes when its cube-local position $q$ satisfies $|q_y-\epsilon a_y|\leq.02$, $|q_x|\leq.075$, and $|q_z|\leq.075$. Its contribution is $[-\epsilon(R_c^\top f_c)_y]_+$, where $f_c$ is force on the cube. $F_h^{\mathrm{full}}$ instead permits $|q_x|\leq a_x+.01$ and $|q_z|\leq a_z+.01$. These load measurements return zero if the history is shorter than $.08$ s.

Let $p_H,e_H,l_H,r_H$ denote the handle centre, axial direction, half-length, and radius. Its sampled surface gap to the cube is
\[
 g_H=\left[\min_{t\in\{-1,-.5,0,.5,1\}}
 \left\|\left[\left|R_c^\top(p_H+t l_H e_H-p_c)\right|-a\right]_+\right\|_2-r_H\right]_+.
\]
For door centre $p_D$, normal $n_D$, and half-thickness $a_D$, define
\[
 g_D=\left[|(p_c-p_D)^\top n_D|-\sum_j|(R_c^\top n_D)_j|a_j-a_D\right]_+.
\]
$F_H^\downarrow$ is the minimum summed downward cube force on the handle over $.08$ s; $F_D$ is the minimum summed positive cube--panel normal load over the same window. Both return zero for shorter histories. The handle and door angles are $\theta_H$ and $\theta_D$.

\begin{enumerate}
\item \textbf{Approach the cube.}
\[
\begin{split}
 r_1=\operatorname{col}\bigl(&N_{.20,.04}(e_{\mathrm{reach}};0),\ L_{.4}(v_f;\cos30^\circ),\\
 &N_{(.18,.20),(.04,.10)}(b;(.25,0)),\ r_m^{\mathrm{c}},\ U_{.5}(v;.25),\ L_{.3}(u;\cos25^\circ)\bigr).
\end{split}
\]
The motion block accepts $m_p\leq.03$ and $m_R\leq8^\circ$; these are also candidate constraints.
\item \textbf{Grasp opposing side faces.}
\[
 r_2=\operatorname{col}\bigl(N_{.15,.02}(e_{\mathrm{grasp}};0),\ L_5(F_h;5),\ r_m^{\mathrm{c}},\ U_{.5}(v;.25)\bigr).
\]
Each palm need only sustain $3$ N for acceptance. The motion override is unchanged.
\item \textbf{Lift clear of the stand.}
\[
 r_3=\operatorname{col}\bigl(L_{.12}(h_s;.08),\ L_5(F_h;5),\ U_5(F_s;0),\ L_{.3}(u;\cos30^\circ)\bigr).
\]
Acceptance permits $F_s\leq.1$ and requires each palm load to be at least $3$ N.
\item \textbf{Carry to the handle.}
\[
\begin{split}
 r_4=\operatorname{col}\bigl(&U_{.20}(g_H;.02),\ (\sqrt{.5}\,N_{.20,.04}(e_{\mathrm{grasp}};0))^{\mathrm{g}},\\
 &L_{.20}(\underline h_c;.55)^{\mathrm{c}},\ L_5(F_h;5),\ U_5(\overline F_s;0)^{\mathrm{c}},\ L_{.3}(u;\cos30^\circ)\bigr).
\end{split}
\]
The palm-load override remains $3$ N per hand; the stand-load override and constraint permit $\overline F_s\leq.1$.
\item \textbf{Depress the handle with the cube.}
\[
\begin{split}
 r_5=\operatorname{col}\bigl(&L_{.35}(\theta_H;.50),\ L_8(F_H^\downarrow;8),\ U_{.10}(g_H;0),\\
 &U_{.15}(\theta_D;.04),\ L_{.20}(h_c;.15),\ L_5(F_h;5)^{\mathrm{c}},\ L_{.3}(u;\cos30^\circ)\bigr).
\end{split}
\]
Acceptance permits $\theta_D\leq.06$ and requires $F_H^\downarrow\geq3$ and both palm loads at least $3$ N. The palm-load condition is also a candidate constraint.
\item \textbf{Push the door open.}
\[
\begin{split}
 r_6=\operatorname{col}\bigl(&L_{.5}(\theta_D;1),\ L_5(F_D;5),\ U_{.10}(g_D;0),\\
 &L_{.20}(h_c;.15),\ L_5(F_h^{\mathrm{full}};5),\ L_{.3}(u;\cos30^\circ)\bigr).
\end{split}
\]
Acceptance requires $\theta_D\geq.95$, $F_D\geq3$, and each full-face palm load at least $3$ N.
\end{enumerate}

\subsubsection{Hurdle traversal}
\label{app:hurdle_math}
The fixed hurdle has near and far faces $x_n,x_f$, top height $z_h$, and half-width $w$. The right leg leads and the left follows. For collision group $G$, let $x_G,z_G$ be the minimum world $x,z$ coordinates of its collision surface. Mesh vertices and sphere extrema provide these measurements. Write $T_G$ and $F_G$ for its upward hurdle-top load and positive floor normal load. A top contact is counted only if its height is at least $z_h-.015$ and its contact normal has vertical magnitude greater than $.7$. Let $A_n$ be the summed positive normal load on hands, wrists, and elbows at contacts with $x\leq x_n+.015$ and $z<z_h-.015$. Define
\[
 C_b=U_{.20}(|y_{\mathrm{root}}|;b),\qquad
 K=U_{20}(A_n;2),\qquad
 B(\xi)=\left[\cos35^\circ-\min_{s\in\xi}(R_{\mathrm{torso}}(s))_{zz}\right]_+.
\]
$K$ is always guidance-only. Let $u=(R_{\mathrm{torso}})_{zz}$, $l=(R_{\mathrm{torso}})_{xz}$, $f_y=(R_{\mathrm{torso}})_{yx}$, and $W$ be total model mass times gravitational acceleration. Subscripts $R,L$ denote right and left feet; $\mathrm{leg}R,\mathrm{leg}L$ include collision bodies descending from the respective knee. $A,P,\mathcal R$ denote arms, pelvis, and entire robot.

\paragraph{Placement and clearance measurements.}
For each palm's collision bounding-box centre $b_h$, define
\[
 e_h=\operatorname{col}\left(
 b_{h,x}-\operatorname{clip}(b_{h,x},x_n+.025,x_f-.025),\,
 b_{h,y}-\operatorname{clip}(b_{h,y},-w_p,w_p),\,
 z_h^{\mathrm{hand}}-z_h\right),
\]
where $w_p=\min(.35,w-.08)$ and $z_h^{\mathrm{hand}}$ is the hand's lowest collision point. Stack both hands into $e_H$; $n_H$ stacks downward palm-normal components. Let $g_G$ be the minimum vertex distance of group $G$ to the top rectangle $[x_n+.02,x_f-.02]\times[-w+.08,w-.08]\times\{z_h\}$. The pelvis error $e_P$ comprises its bounding-box centre's signed distance from $[x_n+.02,x_f-.02]$ in $x$ and its lowest collision height minus $z_h$.

Define right-foot and trailing-leg clearance errors
\[
 a_R=\begin{cases}[z_h+.02-z_R]_+,&x_R<x_f+.02,\\0,&\text{otherwise},\end{cases}
 \qquad
 a_L=\left[z_h+.02-\min_{\substack{p\in\mathcal P_{\mathrm{leg}L}\\p_x<x_f+.02}}p_z\right]_+,
\]
with $a_L=0$ when no vertices remain. Let $d_L=[.05-\min(d_{\mathrm{knee}},d_{\mathrm{ankle}})]_+$, where each $d$ is the left joint's displacement from the pelvis projected onto the torso's forward axis. Finally, $v=\|\dot p_{\mathrm{root}}\|_2$ and $\omega=\|\omega_{\mathrm{root}}\|_2$.

\begin{enumerate}
\item \textbf{Both palms on top.}
\[
 r_1=\operatorname{col}\bigl(C_{.36},U_{.20}(|e_H|;.035),L_1(n_H;\cos40^\circ),L_2(T_{H_L};2),L_2(T_{H_R};2)\bigr).
\]
\item \textbf{Right leg on top for support.}
\[
\begin{split}
 r_2=\operatorname{col}\bigl(&C_{.36},U_{.20}(g_{\mathrm{leg}R};0),
 L_{.20}((p_{\mathrm{knee}R,x},p_{\mathrm{knee}R,z});(x_n+.04,z_h+.08))^{\mathrm{g}},\\
 &L_{10}(T_{\mathrm{leg}R};10),L_{20}(\min_{s\in\xi}T_A(s);5)^{\mathrm{c}},
 L_{.20W}(T_A;.20W)^{\mathrm{g}},K^{\mathrm{g}}\bigr).
\end{split}
\]
The leg-gap block accepts $g_{\mathrm{leg}R}\leq.025$.
\item \textbf{Right foot on top.}
\[
 r_3=\operatorname{col}\bigl(L_{.20}(x_R;x_n+.02),L_{.20}(z_R;z_h-.01),L_5(T_R;5),L_{20}(F_L;5),K^{\mathrm{g}}\bigr).
\]
\item \textbf{Lean and sweep the right foot across.}
\[
 r_4=\operatorname{col}\bigl(C_{.36},L_{.30}(l;\sin30^\circ),U_{.20}(a_R;0),L_{.20}(x_R;x_f+.02),L_{20}(F_L;5),L_{10}(T_{\mathcal R};10),K^{\mathrm{g}}\bigr).
\]
\item \textbf{Lift the left foot into a supported straddle.}
\[
\begin{split}
 r_5=\operatorname{col}\bigl(&C_{.35},L_{.30}(l;\sin10^\circ),L_{.20}(x_R;x_f+.02),L_{.20}(z_L;.08),\\
 &U_W(F_L;0)^{\mathrm{g}},L_{.20}(x_{\mathrm{root}};x_n+.02),L_{20}(T_{\mathcal R};10),K^{\mathrm{g}}\bigr).
\end{split}
\]
\item \textbf{Settle into an upright supported straddle.}
\[
\begin{split}
 r_6=\operatorname{col}\bigl(&C_{.36},L_{.30}(u;\cos15^\circ),L_{.30}(f_y;\cos30^\circ),e_P/.20,\\
 &L_{.20}(x_R;x_f+.02),U_{.30}(p_{\mathrm{ankle}R,z}-p_{P,z};-.10),L_{.20}(z_L;.08),\\
 &U_{.50}(v;.20),U_1(\omega;.50),L_{10}(T_P;10),K^{\mathrm{g}}\bigr).
\end{split}
\]
The pelvis-placement block accepts $\|e_P\|_\infty\leq.025$.
\item \textbf{Left foot on the ledge.}
\[
 r_7=\operatorname{col}\bigl(g_L/.20,d_L/.20,L_{20}(T_L;10),L_{.20}(z_L;z_h-.01)^{\mathrm{g}},B(\xi)/.20\bigr).
\]
Acceptance requires $g_L/.20\leq.025$, $d_L=0$, $T_L\geq10$, and terminal $u\geq\cos35^\circ$. The historical balance penalty may remain nonzero.
\item \textbf{Left leg clears the far edge.}
\[
 r_8=\operatorname{col}\bigl(L_{.20}(\min_{s\in\xi}x_L(s);x_n-.04)^{\mathrm{c}},L_{.20}(x_L;x_f+.03),a_L/.20,B(\xi)/.20\bigr).
\]
The forward residual accepts $L_{.20}(x_L;x_f+.03)\leq.05$; the remaining blocks use their default zero tests.
\item \textbf{Both feet on the far-side floor.}
\[
\begin{split}
 r_9=\operatorname{col}\bigl(&L_{.20}(\min_{s\in\xi}x_L(s);x_f+.01)^{\mathrm{c}},\\
 &\{L_{.20}(x_i;x_f+.03),U_{.20}(z_i;.04),L_{10}(F_i;5)\}_{i\in\{L,R\}},(B(\xi)/.20)^{\mathrm{g}}\bigr).
\end{split}
\]
\item \textbf{Stand and settle.}
Let $p^\star=(x_f+.50,y_{\mathrm{root}},z_{\mathrm{nom}})$ and let $e_{\mathrm{rot}}$ be the quaternion rotation difference from the identity orientation. With nominal joint positions $q_{\mathrm{nom}}$, define
\[
 e_{\mathrm{stand}}=\operatorname{col}\bigl((p_{\mathrm{root}}-p^\star)/.20,e_{\mathrm{rot}}/.5,q_{\mathrm{joints}}-q_{\mathrm{nom}}\bigr),
 \qquad
 r_{10}=\operatorname{col}\bigl(e_{\mathrm{stand}}^{\mathrm{g}},[\overline v-.15]_+\bigr).
\]
Here temporal extrema use the final $.5$ s. Acceptance requires at least $.5$ s of history, $\underline u\geq\cos15^\circ$, $\overline v\leq.15$, $\min_{i\in\{L,R\}}\underline F_i\geq5$, and $\min_{s}(x_{\mathcal R}(s)-x_f)\geq.03$ over that window.
\end{enumerate}

\subsubsection{Carrying a box through points}
\label{app:carry_math}
{This is the task of the sanity check in Appendix~\ref{app:sanity_check}. Let $p_b$ be the box centre and $q_k$ the $k$-th point of the path, in a world frame with $x$ forward, $y$ to the robot's left, and $z$ up. The box starts on the post at $(.306,0,.891)$. With $t_{\mathrm{end}}$ the time of the final frame of $\xi$, the time since the box first touched the floor is
\[
 \tau_{\mathrm{floor}}(\xi)=\Bigl[t_{\mathrm{end}}-\min\{t(s):s\in\xi,\ \text{the box touches the floor at }s\}\Bigr]_+,
\]
with $\tau_{\mathrm{floor}}=0$ if the box never touches the floor. Every effect $k=1,\ldots,10$ has the same two blocks,
\[
 r_k=\operatorname{col}\bigl((p_b-q_k)/.10,\ \tau_{\mathrm{floor}}^{\mathrm{c}}\bigr).
\]
The point block accepts $\|p_b-q_k\|_2\leq.05$, and the floor block requires $\tau_{\mathrm{floor}}=0$. Nothing else is measured. Table~\ref{tab:carry_points} lists the points of the three paths.}

\begin{table}[h]

\centering
\scriptsize
\caption{\textbf{Points of the three generated paths} for the sanity check, box-centre positions in metres.}
\label{tab:carry_points}
\begin{tabular}{@{}cccc@{}}
\toprule
Point & Path 1 & Path 2 & Path 3 \\
\midrule
1 & (0.300, $-$0.050, 0.980) & (0.300, $-$0.050, 0.980) & (0.300, 0.050, 0.980) \\
2 & (0.303, $-$0.350, 0.985) & (0.425, $-$0.323, 0.940) & (0.403, 0.332, 0.963) \\
3 & (0.458, $-$0.607, 0.940) & (0.462, $-$0.621, 1.007) & (0.623, 0.535, 1.001) \\
4 & (0.408, $-$0.903, 0.940) & (0.351, $-$0.899, 0.957) & (0.665, 0.832, 1.068) \\
5 & (0.488, $-$1.192, 0.940) & (0.257, $-$1.184, 0.986) & (0.901, 1.017, 1.129) \\
6 & (0.464, $-$1.491, 0.940) & (0.318, $-$1.478, 1.032) & (1.182, 1.124, 1.073) \\
7 & (0.561, $-$1.775, 1.014) & (0.446, $-$1.749, 1.018) & (1.355, 1.369, 1.067) \\
8 & (0.554, $-$2.075, 1.058) & (0.549, $-$2.031, 1.069) & (1.586, 1.560, 0.987) \\
9 & (0.338, $-$2.283, 1.102) & (0.477, $-$2.322, 1.124) & (1.747, 1.813, 0.940) \\
10 & (0.042, $-$2.333, 1.090) & (0.308, $-$2.570, 1.058) & (1.803, 2.108, 0.940) \\
\bottomrule
\end{tabular}
\end{table}

\FloatBarrier
\section{Tips for practitioners}
\label{app:tips}

\paragraph{Begin with an intuitive score.}
  Highly constrained scores are not necessarily better scores. We first encode the obvious physical outcome and inspect the behavior produced by the search. We add terms only to address failure modes observed in the resulting rollouts, rather than trying to anticipate every possible failure in advance.

  \paragraph{Define the task one physical effect at a time.}
  The terminal rollout of one accepted effect often reveals how the next effect should be specified. It may expose the relevant contact configuration, show which movement is currently blocked, or indicate that an intermediate physical effect is required. We therefore found it more practical to construct long tasks stage-wise than to define every effect in advance.
  
  \paragraph{Inspect the motion, not only the scalar score.}
  A lower physical-effect score does not always indicate useful progress, and a temporary score increase does not always indicate failure. The robot may need to make a short adjustment before further progress becomes possible, while a search can also exploit a missing condition and reduce the score through an unintended movement. Video replay remains the most useful diagnostic for distinguishing these cases.

  \paragraph{Avoid ending a stage in uncontrolled motion.}
  Dynamic or unexpected terminal states are difficult to recover from during later effects. When this occurs, the next score should not merely force the robot forward from an unsuitable state. It is usually better to revise the preceding effect to require settling, support, or bounded velocity before acceptance.

  \paragraph{Match the horizon to the physical response.}
  Large body movements and heavy-object displacement may not develop within the same horizon as a small contact adjustment. Increasing the prediction horizon gives the simulator time to reveal the outcome of a candidate, while increasing the execution horizon can reduce the number of replanning decisions. Both can also make an incorrect candidate more costly. We therefore treat prediction and execution horizons as task-stage parameters.

\FloatBarrier
\end{document}

%% file: math_commands.tex
\usepackage{amsmath,amsfonts,bm}

\def\eqref#1{equation~\ref{#1}}

\def\1{\bm{1}}

\DeclareMathAlphabet{\mathsfit}{\encodingdefault}{\sfdefault}{m}{sl}
\SetMathAlphabet{\mathsfit}{bold}{\encodingdefault}{\sfdefault}{bx}{n}



%% file: iclr2027_conference.bib
@InProceedings{dan2025xsim,
  title = 	 {X-Sim: Cross-Embodiment Learning via Real-to-Sim-to-Real},
  author =       {Dan, Prithwish and Kedia, Kushal and Chao, Angela and Duan, Edward and Pace, Maximus Adrian and Ma, Wei-Chiu and Choudhury, Sanjiban},
  booktitle = 	 {Proceedings of The 9th Conference on Robot Learning},
  pages = 	 {816--833},
  year = 	 {2025},
  editor = 	 {Lim, Joseph and Song, Shuran and Park, Hae-Won},
  volume = 	 {305},
  series = 	 {Proceedings of Machine Learning Research},
  month = 	 {27--30 Sep},
  publisher =    {PMLR},
  url = 	 {https://proceedings.mlr.press/v305/dan25a.html}
}

@misc{ranawaka2026simfoundrymodularautomatedscene,
      title={SimFoundry: Modular and Automated Scene Generation for Policy Learning and Evaluation}, 
      author={Nadun Ranawaka and Josiah Wong and Wei-Lin Pai and Wei-Teng Chu and Tianyuan Dai and Masoud Moghani and Hang Yin and Yunfan Jiang and Wesley Durbano and Brandon Huynh and Yu Fang and Danfei Xu and Ruohan Zhang and Li Fei-Fei and Linxi Fan and Bowen Wen and Ajay Mandlekar and Yuke Zhu},
      year={2026},
      eprint={2606.28276},
      archivePrefix={arXiv},
      primaryClass={cs.RO},
      url={https://arxiv.org/abs/2606.28276}, 
}

@inproceedings{
DrS,
title={DrS: Learning Reusable Dense Rewards for Multi-Stage Tasks},
author={Tongzhou Mu and Minghua Liu and Hao Su},
booktitle={The Twelfth International Conference on Learning Representations},
year={2024},
url={https://openreview.net/forum?id=6CZ50WgfCG}
}

@article{chen2025gmt,
title={GMT: General Motion Tracking for Humanoid Whole-Body Control},
author={Chen, Zixuan and Ji, Mazeyu and Cheng, Xuxin and Peng, Xuanbin and Peng, Xue Bin and Wang, Xiaolong},
journal={arXiv:2506.14770},
year={2025}
}

@article{
liao2025beyondmimicmotiontrackingversatile,
author = {Qiayuan Liao  and Takara E. Truong  and Xiaoyu Huang  and Yuman Gao  and Guy Tevet  and Koushil Sreenath  and C. Karen Liu },
title = {BeyondMimic: From motion tracking to versatile humanoid control via guided diffusion},
journal = {Science Robotics},
volume = {11},
number = {117},
pages = {eadx8924},
year = {2026},
doi = {10.1126/scirobotics.adx8924},
URL = {https://www.science.org/doi/abs/10.1126/scirobotics.adx8924},
eprint = {https://www.science.org/doi/pdf/10.1126/scirobotics.adx8924}}

@misc{feng2026autonomousmathematicsresearch,
      title={Towards Autonomous Mathematics Research}, 
      author={Tony Feng and Trieu H. Trinh and Garrett Bingham and Dawsen Hwang and Yuri Chervonyi and Junehyuk Jung and Joonkyung Lee and Carlo Pagano and Sang-hyun Kim and Federico Pasqualotto and Sergei Gukov and Jonathan N. Lee and Junsu Kim and Kaiying Hou and Golnaz Ghiasi and Yi Tay and YaGuang Li and Chenkai Kuang and Yuan Liu and Hanzhao Lin and Evan Zheran Liu and Nigamaa Nayakanti and Xiaomeng Yang and Heng-Tze Cheng and Demis Hassabis and Koray Kavukcuoglu and Quoc V. Le and Thang Luong},
      year={2026},
      eprint={2602.10177},
      archivePrefix={arXiv},
      primaryClass={cs.LG},
      url={https://arxiv.org/abs/2602.10177}, 
}

@inproceedings{
jing2025humanoidgen,
title={HumanoidGen: Data Generation for Bimanual Dexterous Manipulation via {LLM} Reasoning},
author={Zhi Jing and Siyuan Yang and Jicong Ao and Ting Xiao and Yu-Gang Jiang and Chenjia Bai},
booktitle={The Thirty-ninth Annual Conference on Neural Information Processing Systems},
year={2025},
url={https://openreview.net/forum?id=Mk9ykil8eP}
}

@misc{taouil2026motiondisco,
      title={{MotionDisco}: Motion Discovery for Extreme Humanoid Loco-Manipulation}, 
      author={Ilyass Taouil and Michal Ciebelski and Shafeef Omar and Haizhou Zhao and Angela Dai and Aaron M. Johnson and Majid Khadiv},
      year={2026},
      eprint={2606.06139},
      archivePrefix={arXiv},
      primaryClass={cs.RO},
      url={https://arxiv.org/abs/2606.06139}, 
}

@inproceedings{
tirinzoni2025zeroshot,
title={Zero-Shot Whole-Body Humanoid Control via Behavioral Foundation Models},
author={Andrea Tirinzoni and Ahmed Touati and Jesse Farebrother and Mateusz Guzek and Anssi Kanervisto and Yingchen Xu and Alessandro Lazaric and Matteo Pirotta},
booktitle={The Thirteenth International Conference on Learning Representations},
year={2025},
url={https://openreview.net/forum?id=9sOR0nYLtz}
}

@inproceedings{
li2026bfmzero,
title={{BFM}-Zero: A Promptable Behavioral Foundation Model for Humanoid Control Using Unsupervised Reinforcement Learning},
author={Yitang Li and Zhengyi Luo and Tonghe Zhang and Cunxi Dai and Anssi Kanervisto and Andrea Tirinzoni and Haoyang Weng and Kris Kitani and Mateusz Guzek and Ahmed Touati and Alessandro Lazaric and Matteo Pirotta and Guanya Shi},
booktitle={The Fourteenth International Conference on Learning Representations},
year={2026},
url={https://openreview.net/forum?id=jkhl2oI0g5}
}

@article{FBMEBE,
    title={Maximum Entropy Behavior Exploration for Sim2Real Zero-Shot Reinforcement Learning},
    author={Jiajun Hu and N{\'u}ria Armengol Urp{\'\i} and Jin Cheng and Stelian Coros},
    journal={Reinforcement Learning Journal},
    volume={7},
    year={2026}
}

@inproceedings{
FB,
title={Learning One Representation to Optimize All Rewards},
author={Ahmed Touati and Yann Ollivier},
booktitle={Advances in Neural Information Processing Systems},
editor={A. Beygelzimer and Y. Dauphin and P. Liang and J. Wortman Vaughan},
year={2021},
url={https://openreview.net/forum?id=2a96Bf7Qdrg}
}

@inproceedings{
METRA,
title={{METRA}: Scalable Unsupervised {RL} with Metric-Aware Abstraction},
author={Seohong Park and Oleh Rybkin and Sergey Levine},
booktitle={The Twelfth International Conference on Learning Representations},
year={2024},
url={https://openreview.net/forum?id=c5pwL0Soay}
}

@inproceedings{
HILP,
title={Foundation Policies with Hilbert Representations},
author={Seohong Park and Tobias Kreiman and Sergey Levine},
booktitle={Forty-first International Conference on Machine Learning},
year={2024},
url={https://openreview.net/forum?id=LhNsSaAKub}
}

@article{SONIC,
author = {Zhengyi Luo  and Ye Yuan  and Tingwu Wang  and Chenran Li  and Fernando Castañeda  and Sirui Chen  and Zi-Ang Cao  and Jiefeng Li  and David Minor  and Qingwei Ben  and Jinhyung Park  and David Sami  and Zi Wang  and Xingye Da  and Runyu Ding  and Cyrus Hogg  and Lina Song  and Edy Lim  and Eugene Jeong  and Tairan He  and Haoru Xue  and Wenli Xiao  and Simon Yuen  and Jan Kautz  and Yan Chang  and Umar Iqbal  and Linxi “Jim” Fan  and Yuke Zhu },
title = {SONIC: Supersizing motion tracking for natural humanoid whole-body control},
journal = {Science Robotics},
volume = {11},
number = {117},
pages = {eaed4592},
year = {2026},
doi = {10.1126/scirobotics.aed4592},
URL = {https://www.science.org/doi/abs/10.1126/scirobotics.aed4592},
eprint = {https://www.science.org/doi/pdf/10.1126/scirobotics.aed4592}}

@InProceedings{TWIST,
  title = 	 {TWIST: Teleoperated Whole-Body Imitation System},
  author =       {Ze, Yanjie and Chen, Zixuan and Araujo, Joao Pedro and Cao, Zi-ang and Peng, Xue Bin and Wu, Jiajun and Liu, Karen},
  booktitle = 	 {Proceedings of The 9th Conference on Robot Learning},
  pages = 	 {2143--2154},
  year = 	 {2025},
  editor = 	 {Lim, Joseph and Song, Shuran and Park, Hae-Won},
  volume = 	 {305},
  series = 	 {Proceedings of Machine Learning Research},
  month = 	 {27--30 Sep},
  publisher =    {PMLR},
  url = 	 {https://proceedings.mlr.press/v305/ze25a.html}
}

@misc{PGMT,
      title={PGMT: Perceptive General Motion Tracking for Humanoid Robots}, 
      author={Hongyi Li and Li Peizhuo and Yucheng Tao and Ze Wang and Fangzhou Xu and Jinyi Chen and Yanyan Yuan and Dapeng Jia and Yongbin Jin and Mingfeng Fan and Guillaume Sartoretti and Hongtao Wang},
      year={2026},
      eprint={2609.08511},
      archivePrefix={arXiv},
      primaryClass={cs.RO},
      url={https://arxiv.org/abs/2609.08511}, 
}

@misc{GigaBrainWBC,
      title={GigaBrain-WBC-0.5: A Behavior World Model for Robust Humanoid Whole-Body Tracking with Environment Interaction}, 
      author={Ziyang Cheng and Tianshu Tang and Jinxin Lan and Xinze Chen and Yuhan Gong and Zhichao Liu and Changzhong Wu and Yahao Mao and Zongyan Deng and Mingxuan Ma and Huasen Xi and Yilong Liu and Yutong Wu and Xiaofeng Wang and Borui Zhang and Bingyao Yu and Yang Wang and Yun Ye and Guan Huang and Xiaojie Jin and Zheng Zhu and Jiwen Lu},
      year={2026},
      eprint={2608.18234},
      archivePrefix={arXiv},
      primaryClass={cs.RO},
      url={https://arxiv.org/abs/2608.18234}, 
}

@inproceedings{SPiRL,
title={Accelerating Reinforcement Learning with Learned Skill Priors},
author={Karl Pertsch and Youngwoon Lee and Joseph J. Lim},
booktitle={Conference on Robot Learning (CoRL)},
year={2020},

}

@InProceedings{SkiMo,
  title = 	 {Skill-based Model-based Reinforcement Learning},
  author =       {Shi, Lucy Xiaoyang and Lim, Joseph J and Lee, Youngwoon},
  booktitle = 	 {Proceedings of The 6th Conference on Robot Learning},
  pages = 	 {2262--2272},
  year = 	 {2023},
  editor = 	 {Liu, Karen and Kulic, Dana and Ichnowski, Jeff},
  volume = 	 {205},
  series = 	 {Proceedings of Machine Learning Research},
  month = 	 {14--18 Dec},
  publisher =    {PMLR},
  url = 	 {https://proceedings.mlr.press/v205/shi23a.html}
}

@inproceedings{
OPAL,
title={{\{}OPAL{\}}: Offline Primitive Discovery for Accelerating Offline Reinforcement Learning},
author={Anurag Ajay and Aviral Kumar and Pulkit Agrawal and Sergey Levine and Ofir Nachum},
booktitle={International Conference on Learning Representations},
year={2021},
url={https://openreview.net/forum?id=V69LGwJ0lIN}
}

@InProceedings{OpenVLA,
  title = 	 {OpenVLA: An Open-Source Vision-Language-Action Model},
  author =       {Kim, Moo Jin and Pertsch, Karl and Karamcheti, Siddharth and Xiao, Ted and Balakrishna, Ashwin and Nair, Suraj and Rafailov, Rafael and Foster, Ethan P and Sanketi, Pannag R and Vuong, Quan and Kollar, Thomas and Burchfiel, Benjamin and Tedrake, Russ and Sadigh, Dorsa and Levine, Sergey and Liang, Percy and Finn, Chelsea},
  booktitle = 	 {Proceedings of The 8th Conference on Robot Learning},
  pages = 	 {2679--2713},
  year = 	 {2025},
  editor = 	 {Agrawal, Pulkit and Kroemer, Oliver and Burgard, Wolfram},
  volume = 	 {270},
  series = 	 {Proceedings of Machine Learning Research},
  month = 	 {06--09 Nov},
  publisher =    {PMLR},
  url = 	 {https://proceedings.mlr.press/v270/kim25c.html}
}

@INPROCEEDINGS{ResidualRL,
  author={Johannink, Tobias and Bahl, Shikhar and Nair, Ashvin and Luo, Jianlan and Kumar, Avinash and Loskyll, Matthias and Ojea, Juan Aparicio and Solowjow, Eugen and Levine, Sergey},
  booktitle={2019 International Conference on Robotics and Automation (ICRA)}, 
  title={Residual Reinforcement Learning for Robot Control}, 
  year={2019},
  volume={},
  number={},
  pages={6023-6029},
  doi={10.1109/ICRA.2019.8794127}}

@inproceedings{SUPE,
  title={Leveraging Skills from Unlabeled Prior Data for Efficient Online Exploration},
  author={Wilcoxson, Max and Li, Qiyang and Frans, Kevin and Levine, Sergey},
  booktitle={International Conference on Machine Learning (ICML)},
  year={2025},
}

@inproceedings{murray2026flowdagger,
  title={{FlowDAgger}: Human-in-the-Loop Adaptation of Generative Robot Policies in Latent Space},
  author={Murray, Michael and Chen, Daphne and Bagaria, Simran and Fortier, Dean and Hellebrekers, Tess and Mullins, Galen and Gajarla, Harshavardhan and Mees, Oier and Cakmak, Maya and Kolobov, Andrey},
  booktitle={Proceedings of the 10th Conference on Robot Learning},
  year={2026}
}

@misc{UniSteer,
      title={UniSteer: Unified Noise Steering for Efficient Human-Guided VLA Adaptation}, 
      author={Junjie Lu and Xinyao Qin and Yuhua Jiang and Kaixin Wang and Chuheng Zhang and Bin Liang and Jun Yang and Min Xu and Li Zhao},
      year={2026},
      eprint={2605.10821},
      archivePrefix={arXiv},
      primaryClass={cs.RO},
      url={https://arxiv.org/abs/2605.10821}, 
}

@inproceedings{
FlowReversalSteering,
title={Improving Robotic Generalist Policies via Flow Reversal Steering},
author={Andy Tang and William Chen and Andrew Wagenmaker and Chelsea Finn and Sergey Levine},
booktitle={3rd Workshop on Semantic Reasoning and Goal Understanding in Robotics (RSS 2026)},
year={2026},
url={https://openreview.net/forum?id=lZLHs6P90d}
}

@InProceedings{gSDE,
  title = 	 {Smooth Exploration for Robotic Reinforcement Learning},
  author =       {Raffin, Antonin and Kober, Jens and Stulp, Freek},
  booktitle = 	 {Proceedings of the 5th Conference on Robot Learning},
  pages = 	 {1634--1644},
  year = 	 {2022},
  editor = 	 {Faust, Aleksandra and Hsu, David and Neumann, Gerhard},
  volume = 	 {164},
  series = 	 {Proceedings of Machine Learning Research},
  month = 	 {08--11 Nov},
  publisher =    {PMLR},
  url = 	 {https://proceedings.mlr.press/v164/raffin22a.html}
}

@article{LATTICE,
  title={Latent exploration for reinforcement learning},
  author={Chiappa, Alberto Silvio and Vargas, Alessandro Marin and Huang, Ann Zixiang and Mathis, Alexander},
  journal={Advances in Neural Information Processing Systems (NeurIPS)},
  year={2023}
}

@inproceedings{DEPRL,
  title = {DEP-RL: Embodied Exploration for Reinforcement Learning in Overactuated and Musculoskeletal Systems},
  author = {Schumacher, Pierre and Haeufle, Daniel F.B. and B{\"u}chler, Dieter and Schmitt, Syn and Martius, Georg},
  booktitle = {Proceedings of the Eleventh International Conference on Learning Representations (ICLR)},
  month = may,
  year = {2023},
  doi = {},
  url = {https://openreview.net/forum?id=C-xa_D3oTj6},
  month_numeric = {5}
}

@article{MPPI,
author = {Williams, Grady and Aldrich, Andrew and Theodorou, Evangelos},
year = {2017},
month = {01},
pages = {344-357},
title = {Model Predictive Path Integral Control: From Theory to Parallel Computation},
volume = {40},
journal = {Journal of Guidance, Control, and Dynamics},
doi = {10.2514/1.G001921}
}

@inproceedings{iCEM,
title = {Sample-efficient Cross-Entropy Method for Real-time Planning},
author = {Pinneri, Cristina and Sawant, Shambhuraj and Blaes, Sebastian and Achterhold, Jan and Stueckler, Joerg and Rolinek, Michal and Martius, Georg},
booktitle = {Conference on Robot Learning 2020},
year = {2020},
url = {https://corlconf.github.io/paper_217}
}

@INPROCEEDINGS{DIALMPC,
  author={Xue, Haoru and Pan, Chaoyi and Yi, Zeji and Qu, Guannan and Shi, Guanya},
  booktitle={2025 IEEE International Conference on Robotics and Automation (ICRA)}, 
  title={Full-Order Sampling-Based MPC for Torque-Level Locomotion Control via Diffusion-Style Annealing}, 
  year={2025},
  volume={},
  number={},
  pages={4974-4981},
  doi={10.1109/ICRA55743.2025.11127320}}

@ARTICLE{Wampler1986,
  author={Wampler, Charles W.},
  journal={IEEE Transactions on Systems, Man, and Cybernetics}, 
  title={Manipulator Inverse Kinematic Solutions Based on Vector Formulations and Damped Least-Squares Methods}, 
  year={1986},
  volume={16},
  number={1},
  pages={93-101},
  doi={10.1109/TSMC.1986.289285}}

@article{Khatib1987,
  title={A unified approach for motion and force control of robot manipulators: The operational space formulation},
  author={Oussama Khatib},
  journal={IEEE J. Robotics Autom.},
  year={1987},
  volume={3},
  pages={43-53},
  url={https://api.semanticscholar.org/CorpusID:7687240}
}

@misc{SafeMutations,
      title={Safe Mutations for Deep and Recurrent Neural Networks through Output Gradients}, 
      author={Joel Lehman and Jay Chen and Jeff Clune and Kenneth O. Stanley},
      year={2018},
      eprint={1712.06563},
      archivePrefix={arXiv},
      primaryClass={cs.NE},
      url={https://arxiv.org/abs/1712.06563}, 
}

@inproceedings{PoMS, series={GECCO ’21},
   title={Policy manifold search: exploring the manifold hypothesis for diversity-based neuroevolution},
   url={http://dx.doi.org/10.1145/3449639.3459320},
   DOI={10.1145/3449639.3459320},
   booktitle={Proceedings of the Genetic and Evolutionary Computation Conference},
   publisher={ACM},
   author={Rakicevic, Nemanja and Cully, Antoine and Kormushev, Petar},
   year={2021},
   month=June, pages={901–909},
   collection={GECCO ’21} }

@article{Posa2014,
  author = {Posa, Michael and Cantu, Cecilia and Tedrake, Russ},
  doi = {10.1177/0278364913506757},
  journal = {International Journal of Robotics Research (IJRR)},
  month = jan,
  number = {1},
  pages = {69--81},
  title = {{A Direct Method for Trajectory Optimization of Rigid Bodies Through Contact}},
  url = {http://ijr.sagepub.com/content/33/1/69.short},
  volume = {33},
  year = {2014}
}

@inproceedings{Toussaint2015,
  title={Logic-Geometric Programming: An Optimization-Based Approach to Combined Task and Motion Planning},
  author={Marc Toussaint},
  booktitle={International Joint Conference on Artificial Intelligence},
  year={2015},
  url={https://api.semanticscholar.org/CorpusID:11309330}
}

@misc{OmniContact,
      title={OmniContact: Chaining Meta-Skills via Contact Flow for Generalizable Humanoid Loco-Manipulation}, 
      author={Runyi Yu and Xiaoyi Lin and Ji Ma and Yinhuai Wang and Koukou Luo and Jiahao Ji and Huayi Wang and Wenjia Wang and Runhan Zhang and Ping Tan and Ting Wu and Ruoli Dai and Qifeng Chen and Lei Han},
      year={2026},
      eprint={2606.26201},
      archivePrefix={arXiv},
      primaryClass={cs.RO},
      url={https://arxiv.org/abs/2606.26201}, 
}

@misc{kurtz2024hydrax,
  title={Hydrax: Sampling-based model predictive control on GPU with JAX and MuJoCo MJX},
  author={Kurtz, Vince},
  year={2024},
  note={https://github.com/vincekurtz/hydrax}
}

@article{Rubinstein1999,
  author  = {Rubinstein, Reuven},
  title   = {The Cross-Entropy Method for Combinatorial
             and Continuous Optimization},
  journal = {Methodology and Computing in Applied Probability},
  year    = {1999},
  volume  = {1},
  number  = {2},
  pages   = {127--190},
  doi     = {10.1023/A:1010091220143}
}

@inproceedings{sferrazza2024humanoidbench,
author = {Sferrazza, Carmelo and Huang, Dun-Ming and Lin, Xingyu and Lee, Youngwoon and Abbeel, Pieter},
year = {2024},
month = {07},
pages = {},
title = {HumanoidBench: Simulated Humanoid Benchmark for Whole-Body Locomotion and Manipulation},
doi = {10.15607/RSS.2024.XX.061}
}

@article{OMOMO,
  title={Object Motion Guided Human Motion Synthesis},
  author={Li, Jiaman and Wu, Jiajun and Liu, C Karen},
  journal={ACM Trans. Graph.},
  volume={42},
  number={6},
  year={2023}
}

@inbook{CHOIS,
author = {Li, Jiaman and Clegg, Alexander and Mottaghi, Roozbeh and Wu, Jiajun and Puig, Xavier and Liu, Karen},
year = {2024},
month = {11},
pages = {54-72},
title = {Controllable Human-Object Interaction Synthesis},
isbn = {978-3-031-72939-3},
doi = {10.1007/978-3-031-72940-9_4}
}

@unpublished{hansen2016cocoperformance,
  TITLE = {{COCO: Performance Assessment}},
  AUTHOR = {Hansen, Nikolaus and Auger, Anne and Brockhoff, Dimo and Tusar, Dejan and Tu{\v s}ar, Tea},
  URL = {https://inria.hal.science/hal-01315318},
  NOTE = {ArXiv e-prints, arXiv:1605.03560},
  YEAR = {2016},
  MONTH = May,
  HAL_ID = {hal-01315318},
  HAL_VERSION = {v1},
}

@inproceedings{lum2026play2perfect,
  title={Play2Perfect: What Matters in Dexterous Play Pretraining for Precise Assembly?},
  author={Lum, Tyler Ga Wei and Kedia, Kushal and Liu, Karen and Bohg, Jeannette},
  booktitle={Conference on Robot Learning (CoRL)},
  year={2026}
}

@article{peng2022ase,
   title={ASE: large-scale reusable adversarial skill embeddings for physically simulated characters},
   volume={41},
   ISSN={1557-7368},
   url={http://dx.doi.org/10.1145/3528223.3530110},
   DOI={10.1145/3528223.3530110},
   number={4},
   journal={ACM Transactions on Graphics},
   publisher={Association for Computing Machinery (ACM)},
   author={Peng, Xue Bin and Guo, Yunrong and Halper, Lina and Levine, Sergey and Fidler, Sanja},
   year={2022},
   month=jul, pages={1--17} }

@inproceedings{he2025asap,
  title={{ASAP}: Aligning Simulation and Real-World Physics for Learning Agile Humanoid Whole-Body Skills},
  author={He, Tairan and Gao, Jiawei and Xiao, Wenli and Zhang, Yuanhang and Wang, Zi and Wang, Jiashun and Luo, Zhengyi and He, Guanqi and Sobanbabu, Nikhil and Pan, Chaoyi and Yi, Zeji and Qu, Guannan and Kitani, Kris and Hodgins, Jessica K. and Fan, Linxi and Zhu, Yuke and Liu, Changliu and Shi, Guanya},
  booktitle={Robotics: Science and Systems (RSS)},
  year={2025}
}

@inproceedings{yu2026ssr,
  title={{SSR}: Scaling Surefooted and Symmetric Humanoid Traversal Toward the Open World},
  author={Yu, Ruiqi and Wang, Yiwen and Hao, Yuan and Wu, Jun and Zhu, Qiuguo},
  booktitle={10th Conference on Robot Learning ({CoRL})},
  year={2026},
  url={https://arxiv.org/abs/2605.30770}
}

@inproceedings{todorov2012mujoco,
  author={Todorov, Emanuel and Erez, Tom and Tassa, Yuval},
  booktitle={2012 IEEE/RSJ International Conference on Intelligent Robots and Systems},
  title={{MuJoCo}: A physics engine for model-based control},
  year={2012},
  pages={5026-5033},
  doi={10.1109/IROS.2012.6386109}
}

@article{mittal2025isaaclab,
  title={Isaac Lab: A {GPU}-Accelerated Simulation Framework for Multi-Modal Robot Learning},
  author={Mittal, Mayank and Roth, Pascal and Tigue, James and Richard, Antoine and Zhang, Octi and Du, Peter and Serrano-Mu{\~n}oz, Antonio and Yao, Xinjie and Zurbr{\"u}gg, Ren{\'e} and Rudin, Nikita and others},
  journal={arXiv preprint arXiv:2511.04831},
  year={2025},
  url={https://arxiv.org/abs/2511.04831}
}

@inproceedings{liao2026supermimic,
  title     = {{SuperMimic}: Scaling Humanoid Whole-Body Control with One Differentiable World Model},
  author    = {Liao, Qiayuan and Shao, Yiyang and Long, Junfeng and Zakka, Kevin and Horowitz, Roberto and Abbeel, Pieter and Sreenath, Koushil},
  booktitle = {Conference on Robot Learning (CoRL)},
  year      = {2026},
  url       = {https://arxiv.org/abs/2610.28480}
}
